\documentclass[QA]{sn-jnl}

\jyear{2021}%

\usepackage[numbers,sort&compress]{natbib}
\usepackage{mathtools}

\theoremstyle{thmstyleone}%
\theoremstyle{thmstyletwo}%
\newtheorem{example}{Example}%

\theoremstyle{thmstylethree}%
\newtheorem{definition}{Definition}[subsection]%

\begin{document}
	
	\title[Article Title]{Evaluation of Question Answering Systems: Complexity of judging a natural language}
	
	
	\author[1]{\fnm{Amer} \sur{Farea}}
	\author[1]{Zhen Yang}
	\author[1]{Kien Duong}
	\author[1]{Nadeesha Perera}
	\author*[1]{\fnm{Frank} \sur{Emmert-Streib}}\email{v@bio-complexity.com}
	
	\affil[1]{Predictive Society and Data Analytics Lab, Faculty of Information Technology and Communication Sciences, Tampere University, Finland}

	
	\abstract{Question answering (QA) systems are among the most important and rapidly developing research topics in natural language processing (NLP). A reason, therefore, is that a QA system allows humans to interact more naturally with a machine, e.g., via a virtual assistant or search engine. In the last decades, many QA systems have been proposed to address the requirements of different question-answering tasks. Furthermore, many error scores have been introduced, e.g., based on n-gram matching, word embeddings, or contextual embeddings to measure the performance of a QA system. This survey attempts to provide a systematic overview of the general framework of QA, QA paradigms, benchmark datasets, and assessment techniques for a quantitative evaluation of QA systems. The latter is particularly important because not only is the construction of a QA system complex but also its evaluation. We hypothesize that a reason, therefore, is that the quantitative formalization of human judgment is an open problem. 
	
}


	\maketitle
	
\section{Introduction}\label{sec1}
	
    Natural language processing (NLP) is an important branch of artificial intelligence (AI) concerned with text understanding and text generation. The former subject is studied in the sub-branch \emph{natural language understanding} (NLU) \cite{1,2,3} and the latter in \emph{natural language generation} (NLG) \cite{4,5}. Over the years, both fields, i.e., NLU and NLG developed enormously with an extensive literature which requires nowadays a dedicated discussion of specialized subtasks when presenting approaches or methods thereof despite the fact that a systems understanding of NLP can only be achieved holistically.   

    In this paper, we focus on subtasks of NLP centered around \emph{question answering} (QA). The task of a  QA system is to find an answer (output) in the form of a natural language for a given question (input) usually presented in form of a sentence. While our focus is on QA systems and their quantitative evaluations, we discuss also evaluation scores introduced in related fields. The reason for this is that machine translation (MT), text summarization (TS) and dialogue systems (DS) (also called conversational agents) also have the problem of evaluating various forms of input-output text pairs. Hence, despite the fact that there are considerable differences between question answering, machine translation, text summarization, and dialogue systems several evaluation scores are typically used across those tasks.

    For QA many methods and techniques have been introduced varying from simple rule-based systems to advanced and complex machine learning techniques \cite{8,9,10,11}. In recent years, deep neural network-based approaches to realizing different forms of data-driven representation learning have gained widespread interest due to their competitiveness and flexibility. Question Answering (QA) is widely considered one of the most important tasks of NLP, as it enables humans to interact with machines in a more natural way by either extracting information related to questions from different knowledge sources or by generating the answer without the need for using structural query languages. In other words, QA tries to retrieve or generate answers in the form of a natural language based on a given input, instead of, providing a ranked list of documents as provided by search engines or classic information retrieval systems which require additional steps for checking each document to find useful content \cite{12}.

    Since the 1960s many question-answering systems have been proposed \cite{13,14}. Among the first QA systems were BASEBALL \cite{116} and LUNAR \cite{woods1973progress} which provided information associated with the US baseball league and soil samples from the Apollo lunar exploration. While both approaches provided good results in their corresponding closed-domain applications, a generalization to other domains was impaired. In order to enhance research in open-domain applications, in 1999, the Text REtrieval Conference (TREC) initiated a research area about question answering (TREC-8) in the form of competitions \cite{voorhees1999trec}. Since then there is an ongoing series of campaigns addressing increasingly complex QA problems. As a result, QA systems have evolved to become more relevant to diverse tasks, e.g., generative QA \cite{62,75,163,36}, extractive QA \cite{118,119,120,32}, knowledge-based QA \cite{96,128,129,130,33}, question classification \cite{125,126}, community QA and question answering matching \cite{19,93,94,112,123}.

    The literature about question answering has widely investigated different QA tasks, and many studies have been conducted in this regard according to various criteria such as question types, answer types, knowledge sources, and training strategy techniques \cite{32,33,34,35,36,37,61,80,133,167}. Interestingly, recent studies still focus on how to process a complex question and extract the answer from multiple sources. In contrast, this paper tries to present a broader look at the current state of the literature that is essential for a better understanding of question-answering systems. Specifically, we start by briefly describing the general question-answering framework which allows us to introduce a formal definition of a QA system. For the practical construction of QA systems, we discussed the three main paradigms: (1) Information Retrieval-Based Question Answering (IRQA), (2) Knowledge Base Question Answering (KBQA), and (3) Generative Question Answering (GQA). In order to evaluate a QA system, task-specific benchmark data are needed together with evaluation scores. Due to the fact that there is a large number of such evaluation scores with quite different properties, we introduce a hierarchical taxonomy for such error scores. Furthermore, we discuss important representatives of the main categories of the taxonomy.

    This paper is organized as follows: Section \ref{sec.contributions} briefly indicates the main contributions of this paper. Section \ref{sec.qaframework} introduces a general framework for QA system. Section \ref{sec.paradigms} discusses three different types of the question answering paradigms. Benchmark datasets and evaluation scores are presented in Section \ref{sec.Benchmark}, and different types of quantitative scores are discussed in Section \ref{sec.ces}. In Section \ref{sec.challenges}, we discuss challenges for evaluation scores and in Section \ref{sec.discussion} we provide a general discussion. This paper finishes in Section \ref{sec.conclusion} with a conclusion.

\section{Main contributions of this survey for QA}\label{sec.contributions}

    There are previous review articles about question answering, e.g., \cite{32, 34, 36, 14,33, 35, 54,84,90,171,193,194}, however, each of these studies has investigated a particular type of question answering paradigms and sheds light on some specific requirements of QA tasks (i.e, generative QA, extractive QA, knowledge-based QA, question classification, community QA, or question answering matching) along with some criteria for a specific QA task, such as question types, answer types, knowledge sources, and training strategy techniques. Many studies have also tried to shed light on assessment techniques of NLP in general, and on those used in question-answering systems, e.g., \cite{deriu2021survey, 152, 5,139,140,152}. 

    In contrast to those surveys, our survey is not trying to focus on a specific QA paradigm or on a particular QA system, instead, we attempt to provide a comprehensive overview of the general framework of QA, QA paradigms, benchmark datasets, and assessment techniques for a quantitative evaluation of QA systems. Specifically, our contribution comprises the following topics. 
\begin{enumerate}
    \item Surveying the question answering framework 
    \item Definition of a question-answering system
    \item Question answering paradigms
    \item Benchmark data
    \item Taxonomy of evaluation scores
    \item Discussion of quantitative evaluation scores
\end{enumerate}

    {\bf 1. Surveying the question answering framework:} Question answering has been developed for decades, and there are various designs and approaches for a question answering system. The first goal of the paper is to review all approaches to give a concrete view of the question-answering framework, discussed in Section \ref{sec.qaframework}. We propose a general framework for question answering, which consists of four main components: Question Answering Algorithms, Knowledge Sources, Question Types, and Answer Types. For each component, we explain the different approaches as well as the advantages and drawbacks of those approaches.

    {\bf 2. Definition of a question answering system:} Based on the QA framework, we introduce in Section \ref{sec.defQAsystem} a formal definition of a question answering system which allows summarizing its formal structure abstractly.

    {\bf 3. Question answering paradigms:} In order to provide a clear overview of QA systems, we introduce in Section \ref{sec.paradigms} the three main paradigms of QA systems, i.e., (1) Information Retrieval-Based Question Answering, (2) Knowledge Base Question Answering, and (3) Generative Question Answering and discuss differences among them.

    {\bf 4. Benchmark data:} Since QA systems can be defined for many different and complex tasks with many variations, task-specific datasets are needed for the evaluation of such QA systems. For this reason, we discuss different benchmarking datasets for different QA paradigms. These benchmarking datasets of QA systems are presented in Section \ref{sec.benchdata}.

    {\bf 5. Taxonomy of evaluation scores:} Many different types of evaluation scores have been proposed in the literature, including Untrained Automatic Evaluation Scores (UAES) and Machine-Trained Evaluation Score (MTES), making it hard to distinguish the advantages or drawbacks of those approaches. In Section \ref{sec.errorscores}, we introduce a hierarchical taxonomy of evaluation score classes used for different QA systems.

    {\bf 6. Discussion of quantitative evaluation scores:} For the evaluation score classes of the taxonomy many individual error scores have been introduced. In Section \ref{sec.ces}, we discuss the main representatives for the three major classes of error measures. Specifically, for Simple Untrained Automatic Evaluation Scores (S-UAES) in Section \ref{sec: S-UAES} we discuss exact match (EM), precision, recall, F1-score, error rate (ER), average precision (AP), mean average precision (MAP) and mean reciprocal rank (MRR), for Advanced Untrained Automatic Evaluation Scores (A-UAES) in Section \ref{sec:A-UAES} we discuss  BLEU, NIST, ROUGE and METEOR and for Machine-trained Evaluation Scores (MTES) in Section \ref{sec:MTES} we discuss ADEM, RUBER, RUSE, BLEUERT, BertScore, MaUde, and learning-based composite metrics.

\section{The Question Answering Framework }\label{sec.qaframework}
	
    QA systems have been intensively adopted in many areas like chatbots, search engines, standalone databases, or virtual assistants most notably Alexa, Siri, Cortana, or Google assistance. In general, a QA system accepts questions in the form of text or speech; and tries to parse them into dependency formats \cite{126} to find relationships between their content by looking for matching words or embedded semantics that are located in a context or relevant content \cite{173}. Despite the fact that there are many modern QA systems, which will be discussed below, their underlying framework is similar and is outlined in the following.

\begin{figure}[h]
		\centering
		\includegraphics[width=1\textwidth]{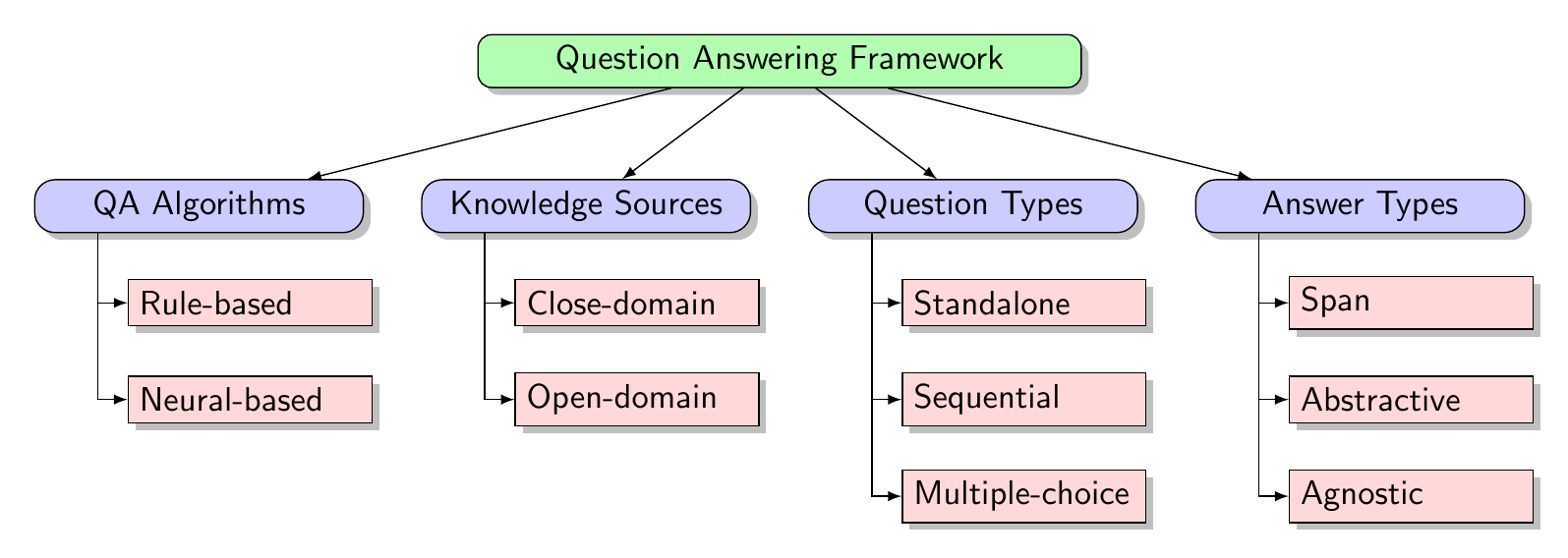}
		\caption{A general framework for a Question Answering system that consists of four main components.  (1) Question Answering Algorithms, (2) Knowledge Sources, (3) Question Types, and (4) and  Answer Types.}
		\label{fig:QAF}
	\end{figure}

	The design of QA approaches may vary from system to system according to the requirements of specific QA tasks. However, a general framework for a QA system, as illustrated in Figure \ref{fig:QAF}, can be structured into the following components: (1) Question Answering Algorithms, (2) Knowledge Sources, (3) Question Types and (4)  Answer Types. These components are described in the following sections.
	
	\subsection{Question Answering Algorithms}
	
    {\it QA algorithms} are at the heart of question answering systems, where they play a vital role in manipulating the given context (i.e., passage, document) and classifying the question/query types of the QA system, and identifying the predicted answer types accordingly. Such algorithms can be classified into {\it rule-based} (heuristic), {\it retrieval-based}, and {\it sequence-to-sequence }techniques. 
 
    {\it Rule-based} techniques can be further subdivided into {\it template-based} \cite{127,132,133}, {\it syntax-based} \cite{78,79,92} and {\it semantic-based} rules \cite{100,105,115}. To identify the answer or question type, traditional methods of QA need to perform some of these rule-based approaches in order to detect the semantics or syntactical parsing of a given input and then produce the required output by converting the in-between representation into a natural language content. Unfortunately, {\it rule-based} techniques are somewhat cumbersome because they usually require large efforts to build handcrafted features, which is considered to be expensive, time-consuming, tedious, and requires an extensive experience in the related domain. In addition, the heuristic technique pipeline may compose many different components, which may affect the quality of the generalization of the selected model. 
 
    Unlike {\it rule-based} approaches, {\it retrieval-based} techniques are using hybrid rules and feature-based classifiers, which can employ plenty of training data and provide an elegant, comprehensive, and trustworthy method to exploit the informative entities of a given input and optimize the content selection along with formulated task (i.e., answer extraction, or question generation) at the same time. A retrieval-based QA system contains two components: intent classifier, and response selector. The intent classifier consists of multiple natural language understanding (NLU) techniques to classify the type of input, understand the intent of the input, and extract the informative entities in the input. Then, the information extracted from the intent classifier is fed to the response selector to query the most appropriate answer from the KB. The techniques for the response selector can be statistical approaches \cite{195} or neural-network approaches \cite{196,197}. Importantly, {\it retrieval-based} techniques do not require handcrafted features; but require instead, large amounts of data to be trained. Also, the obtained results cannot be easily interpreted.
	
    For improving the quality of question answering systems and to free researchers from tedious feature engineering, many neural-based techniques have proposed complex pipelines with deep neural network architectures to formulate the numerous tasks of QA as a {\it sequence-to-sequence} problem. This design boosts the flexibility of a model to handle diverse tasks ranging from synthetic question answering to language modeling \cite{117, 118,120}. Examples for such architectures include {\it encoders and decoders} models \cite{8,82}, {\it attention} \cite{10,20,21,27,47,126,170}, {\it transformer-based} and {\it pre-trained} models \cite{8,58,72, 73,75,122,123,129,146,147,148,162}. The goal behind each model is to encode the relationship between the question and given context,  handle the long-range dependencies in the answer processing modules, and capture the concepts and hidden patterns of hierarchical features in a comprehensive way, to produce an accurate response. 
	
	Despite most of the above-mentioned techniques being found to be more robust 'in somehow' compared to the traditional techniques, such complex techniques require huge datasets along with vast computational resources to be trained. Moreover, a lot of effort is required to fine-tune the model structure to a particular dataset, since there is no single model that can fit efficiently to all QA tasks (each dataset may require different hyperparameters tuning), and the model that performs well for some specific-task dataset may fail to get outstanding result on different tasks or datasets \cite{49}.

\subsection{Knowledge Source}
	
	A {\it knowledge source} in general is either a closed or an open repository that stores data related to a QA system in a structured, unstructured or semi-structured format. Hence, it is the most general source of information that includes a knowledge base. Despite knowledge bases are often in the form of text, due to the growth of signal processing, text KBs are also complemented by audio data \cite{191, 192} and visual data \cite{193, 194}.

\subsubsection{Closed-domain knowledge source}

    A closed-domain could be a single document-based or a compatible and interpretable knowledge-base (KB) based on an entity-linking technique that relies on assigning a unique identity to each entity mentioned in the corpus \cite{157}. In other words, a closed-domain can refer to a single specific knowledge domain, in which the correct answer for an associated question is supposed to be part of a specific document. Closed-domain makes it easy to evaluate a predicted answer against the golden answers; and find the most accurate evaluation for the identical answer at a specific position, especially when the start and end indices of entities are included as labels in the given dataset. For example, in the sentence "[START] Frank [END] was born in [START] 1980 [END].", the golden answers for two questions  "Who was born in 1980?" and "When was Frank born?" are labeled with [START] and [END] indices.

    Many KB datasets \cite{15, 16, 24, 25, 26, 27, 28, 29, 30, 38,39,40,41,42,43,52} have been proposed to tackle the requirements of different applications by providing specific knowledge such as medical, temporal, educational, geospatial to name just a few. The content of a KB is frequently collected from Wikipedia or web pages \cite{158} either as a {\it graph-based} base with entities and edges ({\it subject, relation, object}) or as a {\it logic-based} base, such as  {\it compositional semantic parsing} with labeled entities of precise meaning.

\begin{figure}[h]
		\centering
		\includegraphics[width=0.4\textwidth]{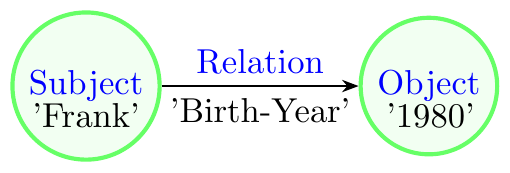}
		\caption{An illustration of a simple knowledge-base with graph-based representation, which uses edges to represent the relationships that exist within a dataset as a collection of nodes (e.g., subjects or objects), in order to predict the answer (object).}
		\label{fig:KB}
	\end{figure}

    {\bf Graph-based:} The {\it graph-based} model uses edges to represent {\it relationships} that link the data items in a dataset to a collection of nodes (e.g., {\it subjects} or {\it objects}) as illustrated in Figure \ref{fig:KB}. Such relations in graph-based models can be used to predict the answer (object) for a question (subject), e.g., "Who was born in 1980?" or "When was Frank born?" by computing the similarity between the encoded question and each possible relation.
	
	The transformers such as BERT \cite{58} or GPT \cite{3, 165} have also been used for a variety of NLP tasks including question answering. For example, BERT has been used to detect the entity and the relation over the question and context, to predict the answer, by using separated trained vectors that represent each possible entity-relation. It starts with tokenizing the given sentences into a vocabulary, normalizes the output of this process, and adds padding when needed. Then it uses the [CLS] and [SEP] tokens to encode a sequence of question and context, and produces the embedding vectors with position and sequence-type information \cite{159}. In the entity span detection step, the embedding vectors are fed into the transformer to detect and predict the start and the end positions of each entity span in the context based on the obtained probability of the start position of the entity span as follows:
	
	\begin{equation}
		P(t=START\vert{x_{0}...x_{T}}) =\frac{e^{h_{i}.w_{START}}}{\sum_{j=0}^{T}e^{h_{j}.w_{START}}}
	\end{equation}
	
	where \(x_{0}...x_{T}\) is the input sequence, \(h_{i}\) is a contextualized   vector feature for timestamp \(i\), and \(w_{START}\) is a parameter vector for the start position classifier. Similarly, the prediction of end positions can be computed using a \(w_{END}\) parameter vector. Then the setting of the transformer model is adjusted to predict the relation \(r\) over the question. The probability of the relation \(R_{i}\) is given by the following formula:
	
	\begin{equation}
		P(r=R_{i}\vert{x_{0}...x_{T}}) =\frac{e^{h_{CLS}.w_{R_{i}}}}{\sum_{k=0}^{N_{R}}e^{h_{CLS}.w_{R_{k}}}}
	\end{equation}
	
	where \(h_{CLS} = h_{1}\) is the contextualized vector feature for the beginning token of the question, \(w_{R_{i}}\) is the parameter vector for relation \(R_{i}\), and \(N_{R}\) is the total number of relation pairs. The entity-relation pairs are created from outputs of the entity span detection and relation detection, re-ranked, and then exploited to query an answer from a knowledge graph.  
	
    {\bf Logic-based:} On another hand, {\it knowledge-base} datasets may contain different structure, e.g., it may contain question-answer-pairs in a logical format, which require  {\it semantic parsing algorithms}, to perform the necessary parsing for such data in a supervised manner, or it may contain question paired with a semantic-answer, in which the logical form needs to be modeled as a latent variable \cite{160,161}. For example, the question "What countries border Finland?" could be formulated as follows:
 
 \begin{eqnarray}
		(\lambda{x}.(countries(x) \land borders(x, Finland))
 \end{eqnarray}
	
\begin{figure}[h!]
		\centering
		\includegraphics[width=0.7\textwidth]{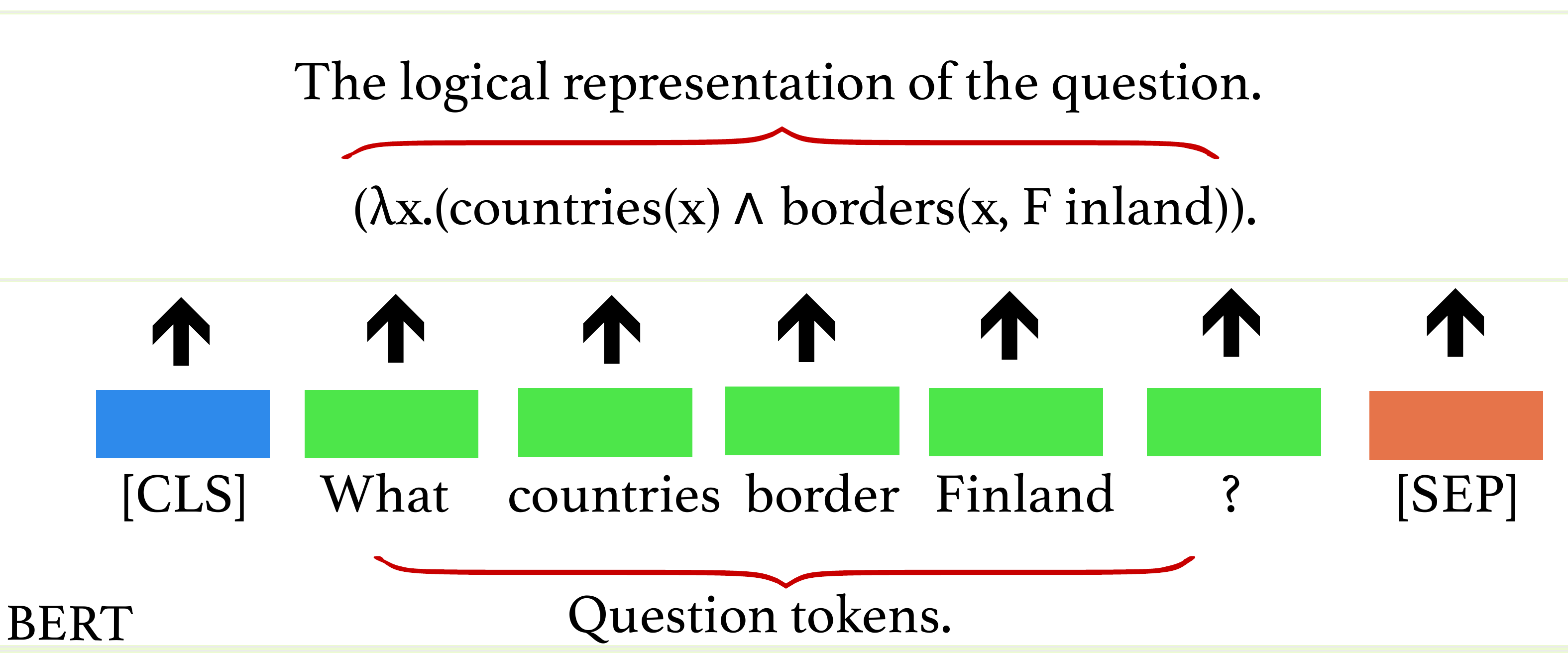}
		\caption{An illustration of a semantic parsing algorithm used by the BERT model over a knowledge-base dataset to formulate a question in a logical representation.}		
		\label{fig:LR}
	\end{figure}

	Besides, the BERT model has been also used with a KB based of {\it semantic parsing} to transform the training tuples of question-answer-pairs into a logical form. This is accomplished by encoding the input sequence; and mapping it to a contextualized vector, followed by a decoding phase to produce a logical representation. An example of this is shown in Figure \ref{fig:LR}.
	
\subsubsection{Open-domain knowledge source}\label{sec.open.domain.ks}
	
\begin{figure}[h]
		\centering
	
			\includegraphics[width=0.9\textwidth]{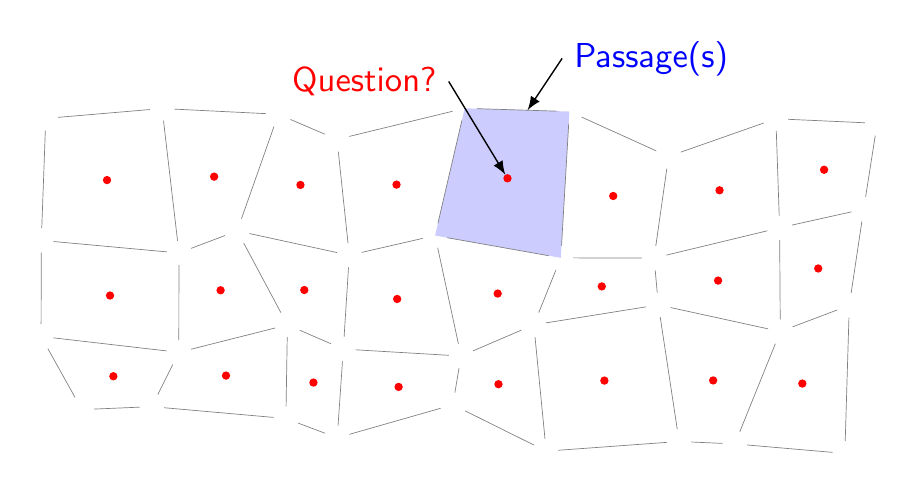}
		\caption{An illustration of the open-domain knowledge source, which illustrates the most relevant document(s)/passage(s) to the open-questions that are being asked. This dataset can be retrieved from various knowledge sources (i.e., {\it texts passage, web documents, knowledge bases, tables, images})}
		\label{fig:od2}
	\end{figure}

	In contrast to closed-domain, an {\it open-domain} knowledge source \cite{17, 18, 19, 20, 21, 22, 23, 31, 44,48,49,50,51,112} is a repository with a huge content that could be collected from various sources (i.e., {\it text passage, books, web documents, knowledge bases, tables, images}), in any possible format, i.e., in the form of tuples ({\it Question, Passage, Answer}), {\it question-answer pairs}, or questions without assuming a given passage, where the answer could be located in some documents that exist in a large collection of documents (e.g., {\it Wikipedia}). In general, open-domain QA datasets can be acquired from an open knowledge source, like the universal ontology, or other crawled web sources, and mostly adopted to answer a general query that does not require specific domain knowledge, as illustrated in Figure \ref{fig:od2}. In other words, an open domain can be referred to as a multiple-document QA, in which the labeled answer for an associated question is not supposed to be located within the specific start and end indices, and also it may not be a part of a particular document (it may be in the same document, in different documents, or not exist at all).

	\subsection{Question and Answer Types}\label{sec.questanswers}
	
    This section addresses Question Types and Answer Types in order to provide an abstract and coherent view of the QA framework in general. It is important to understand Question Types and Answer Types since they are essential elements in all of the QA paradigms, which are discussed in Section \ref{sec.paradigms}

    Determining and categorizing {\it question } and {\it answer types} is an important approach that is needed to obtain the most relevant information with the expected response based on the given input. Thus, QA techniques need to perform different rule-based or neural-based strategies to analyze the input, understand the type of a given structure, and capture the subjects, predicates, patterns, concepts, and conditions that could be embedded in a given content, as well as, in order to detect, extract or validate the answer. Then, the most appropriate answer or generated response could be determined accordingly. Such strategies can follow different paths to achieve this goal, either by using pre-defined patterns to be compared with a given context \cite{127,132}, or by using a semantic representation to automatically discover the patterns of lexical, syntactic, and semantic information embedded in a given context \cite{100,115}, or by parsing the syntactic structure of the given input and extracting named entities (e.g., people, organizations, dates, etc.). The part-of-speech (PoS) tagger, is another strategy to determine the candidate answers  \cite{79,92}. The determined content is then extracted and validated based on a set of heuristics \cite{85,86}. Sequence-to-Sequence or transformer methods are further strategies that could also be used to capture the requirement of the formulated tasks, as mentioned in the previous sections. However, usually, the context with specific topics and information in the closed-domain knowledge source is usually more accurate than those in the open-domain. 

    In the same context, a question type can be {\it standalone, sequential} or {\it multi-choice}. Furthermore a question type can contain various forms of contexts, e.g., {\it factoid, list, causal, hypothetical, long-form} or {\it complex} \cite {84}. 

\begin{itemize}
    \item The most common questions start with {\it wh} (e.g., what, when, where, which, who) and are usually referred to as {\it simple factual} or factoid questions.

    \item Both {\it factoid} and {\it list} questions are having a {\it simple answer} or a short text of continuous tokens, often as a named entity. Such questions do not require deep processing to obtain the answers, they just need to be matched with the answer vectors which are usually represented by a single word or a {\it span of tokens} \cite{173}.
	
    \item The answer to a {\it confirmation} question can be yes or no; and needs to be inferred with quantitative or comparative reasoning. Unfortunately, for some {\it non-factoid} questions like (what-if, what kind, why, how), the answer may appear out of the given context as {\it abstractive answer}. 

    \item The {\it hypothetical} questions (i.e., what-if) usually require information related to an event and depend on a specific context. Therefore, there may not be a specific answer for this type of question and for this reason, the accuracy of the obtained answer could be low. 

    \item The {\it causal} questions require a description of their entities and cannot be answered with just a named entity as the previous question types. This type of question is often containing justified descriptive answers that can span from sentences to paragraphs to an entire document, and it is most commonly used in an open-domain. 

    \item {\it Complex} questions usually consist of several independent questions and often require different types of information that are inferred from multiple sentences, paragraphs, or documents to find an {\it agnostic answer} without the supervision of annotated answers. This type of question is therefore more difficult compared to other types of questions. Once the question is classified \cite{125,122}, then it is possible to determine the related answer using rule-based or neural-based techniques \cite{120,117,121,118,32}. However, when the classification of questions is ambiguous, the system must allow for multiple types of answers. 
\end{itemize}

    Analyzing the question and detecting the answer type, or finding the embeddings of most similar vectors of question and context, is one of the core tasks of QA algorithms. The QA algorithms also facilitate the query formulation during the query processing stage via parsing the passing tokens with precise descriptions in order to identify what kind of answer is supposed to be returned (e.g., paragraph, description, etc). Different strategies have been proposed to deal with such requirements, such as: Regular Expressions (regex), Part-of-Speech (POS), matched n-gram features, finding the first noun phrase after the question headword ‘{\it wh*}’, and  Named Entity Recognition (NER) to identify the given tokens such as human, place, location, entity. The question types classification with different answer types have been investigated in many studies such \cite{125,44,19,48,49,50,51}. The focus detection is meant to capture the question words that are most similar to the answer, while the purpose of the relation extraction is to find the relation that links the entities (i.e., object and subject) in the question to the entity that appears in the context.
	
	A simple question \cite{52} could be represented as bag-of-words or n-grams, and the cosine similarity could be used to find the match between the embedding vector of the candidate answer and the given references \cite{32}. For instance, in \cite{53}, the formulated task for single-fact QA is to predict the candidate subject and candidate relation of a fact $\{subject, relation, object\}$, i.e, the fact "Frank was born in 1980" have the subject "Frank", relation "was born", and object "1980". For such a task, a neural-based model for single-fact QA was proposed to find the most probable relation and subject, by ranking the relations and the entities of the candidate subjects separately. It employs a conditional factoid factorization to infer the probability of the target relation, and then the target subject can be associated with the candidate relation. 
	
    Importantly, modern QA can also deal with complex or narrative questions \cite{54} that may assume different question formats, e.g., hypothesis, confirmation, inferential, cause-effect, etc., which require a different form of answers, i.e., single fact, a span of continuous words, generic paragraph/full-document, focus-generic paragraph, dialog or multiple choices, and are expected to return an answer with a contiguous span of words along with quantitative and comparative reasoning \cite{25,55}.

\subsection{Definition of a question answering system}\label{sec.defQAsystem}

\begin{figure}[t!]
		\centering
		\includegraphics[width=0.9\textwidth]{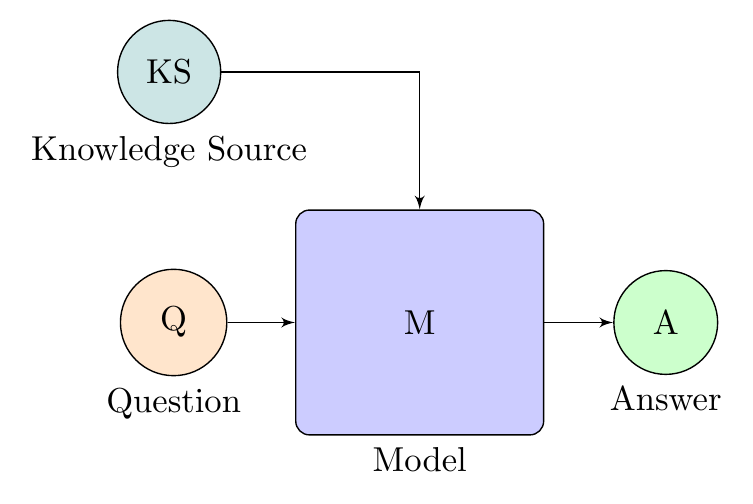}
	\caption{Abstract formulation of a general question answering (QA) system. The QA system is a prediction model (M) where the question (Q) and the knowledge source (KS) form the input of the model (M) and the answer (A) is its output. A knowledge source can be closed or open-domain. Valid question types are either {\it standalone, sequential} or {\it multi-choice} question, and a predicted answer type can be a {\it span, abstractive}, or {\it agnostic}.}
		\label{fig:MM}
	\end{figure}

    A general summary of the description in the previous sections can be provided by the formulation of an abstract QA system. In Figure \ref{fig:MM}, we show a visualization of this abstract model and in the following, we discuss its components in detail. Specifically, in order to define such a QA system one needs to, first, define two components, (i) task and (ii) input data, which will then inform the definition of the prediction model (M).

\begin{definition}[QA task]\label{def.task}
    A QA task is either (i) answer extraction or (ii) response generation.
\end{definition} 

    In Section \ref{sec.qaframework}, we have already seen that for realizing a particular QA system one needs to specify all components of a QA framework (see Figure \ref{fig:QAF}). However, Definition \ref{def.task} refers to a global abstraction level of the overall functioning of a QA system rather than on its individual components. In this respect, we distinguish only two fundamentally different tasks: (i) answer extraction and (ii) response generation. Each of these tasks focuses on a different mechanism of a model (M) for producing an answer, and in Section \ref{sec.paradigms} we will elaborate on different paradigms that can be used therefor and we will see that for answer extraction there are two paradigms, Information Retrieval-based Question Answering (IRQA) and Knowledge Base Question Answering (KBQA) while for response generation the paradigm is Generative Question Answering (GQA). 

    The next component we need to define is the input of a QA system. The input can be described as follows: 
	\begin{equation*}
		\text{Input} = (Q,KS): \left\{
		\begin{array}{ll}
			\text{question,}    & \ Q \  = \{q_{1}, ..., q_{m}\}\\
			\text{knowledge source,} & KS  =\{d_{1}, ..., d_{n}\}\\
		\end{array}
		\right.
		\end{equation*}
    That means the input is a tuple $(Q,KS)$ where $Q$ can consist of questions $q_i$ with $i \in \{1, \dots, m\}$ and a knowledge source $KS$. Valid question types have been discussed in Section \ref{sec.questanswers} and can include standalone, sequential, or multi-choice questions while the knowledge source can contain many documents $d_j$ with $j \in \{1, \dots, n\}$ where each document can contain text passages, web documents, or knowledge bases. The information provided by a knowledge source is formally defined as follows.

\begin{definition}[Knowledge source]\label{def.ks}
    As a knowledge source (KS) we denote the most general source of information storing of data accessible for a QA system. The data can be closed-domain or open-domain and their format can be structured, unstructured, or semi-structured. A knowledge source can also contain a knowledge base (KB).
\end{definition} 

    From Definition \ref{def.ks} one can see that a knowledge source is a very flexible source of information that is not restricted in any way. However, a given task and a specific formulation of a QA problem will in general lead to special cases of a knowledge source. In order to clarify this, we distinguish two representative cases:  

	\begin{eqnarray}
     		\text{Case 1: } KS = \{ D=\{d_{i}\}_{i=1}^{n}, \text{with } q_j, r_k \in D  \label{eqn.case1}\\
		\text { for all } j \in \{1, \dots, m\} \text{ and } k\in \{1, \dots, p\} \} \nonumber
 	\end{eqnarray}
    where $Q=\{q_{j} \}_{j=1}^{m}$ and $R=\{r_{k}\}_{k=1}^{p}$ represent the questions and references (or called gold answers) respectively and $D$ is a set of documents also called a collection. In this case, the questions and gold answers are a subset of the knowledge source, i.e., $Q \subset D$ and $R \subset D$. If the number of questions is the same as the number of answers then $m=p$ and the cardinality of set $Q$ is the same as set $R$, i.e., there are as many answers as questions. However, for multiple or complex answers this is usually not the case which means $m \not = p$. We would like to highlight that the information provided by the gold answer set $R$ is not directly accessible but is sealed within $D$ and the QA system needs to find its elements, i.e., $r_i \in R$.

    A knowledge source with this structure can be used for a QA system that solves either the task of answer extraction or response generation. In Section \ref{sec.paradigms}, we will discuss the QA paradigms, such as Information Retrieval-based Question Answering (IRQA), Knowledge Base Question Answering (KBQA), and Generative Question Answering (GQA) which could all be used for these tasks. However, the task of response generation is commonly considered much harder than answer extraction, and for this reason for a knowledge source with the structure defined in Eqn. \ref{eqn.case1} one would prefer to solve the task via answer extraction.

	\begin{equation}
     		\text{Case 2: } KS = \{ D=\{d_{j}\}_{j=1}^{n} \}
	\end{equation}
    In the second case, the knowledge source does neither contain the questions nor the gold answers. This is the most difficult case because the QA system needs to generate the right responses. Hence, the task of a QA system is answer generation because answer extraction cannot be used. The QA paradigm for this case is Generative Question Answering (GQA) which we will discuss in Section \ref{sec.gqa}.

    We would like to remark that the above cases do not provide an exhaustive list of different knowledge sources but are merely major instances frequently encountered in the literature. Common extensions of the above cases include answer sets allowed to include \emph{no answer}, i.e., $A \cup \text{NULL}$, or KS could include a knowledge base (KB). The latter case would have implications for the QA paradigm allowing to use of Knowledge Base Question Answering (KBQA), discussed in Section \ref{sec.kbqa}.

    This heterogeneity is one of the reasons why QA is such a complex problem with a multitude of individual solutions to suit the corresponding cases, e.g., given by the different forms of the knowledge source (KS).

    Formally, a QA system can now be defined as a prediction model (M) that assumes the following functional form.
	
\begin{definition}[QA system]\label{def.qas}
   	A QA system is a prediction model (M) (either {\it rule-based or neural-based}) realized via a function, $M$, that maps an input space, $(Q; KS)$, to an output space, $A$, corresponding to the predicted answers. Its general definition is given by
	\begin{eqnarray}
	M: (Q; KS) \rightarrow A.
	\end{eqnarray}
    For a particular question $q_i\in Q$ and knowledge source $KS$ a particular answer, $a_i \in A$ is obtained via the mapping
\begin{eqnarray}
    a_i = M(q_i; KS).	
 \end{eqnarray}
\end{definition} 

    From Definition \ref{def.qas}, one can see that the output of a QA system is given by $A$ and can assume the following form:
\begin{equation*}
	\text{Output: Answer}\ A  =\{a_{1}, ..., a_{n}\}\left\{
	\begin{array}{ll}
		 span, & \ 1\leq a_{start}\leq a_{end} \leq N\\
		abstractive \\
		agnostic \\
	\end{array}
	\right.
	\end{equation*}
    Valid answer types of $a_i$ have been discussed in Section \ref{sec.questanswers} and a predicted answer \( {a_{i} \in A}\) (e.g., {\it long answer, span answer or NULL}) can be marked within specific positions (i.e., \(a_{start}\) and \(a_{end}\)) in a given document, while the abstractive answer could be out of a given context. Finally, an agnostic answer can be inferred from multiple sentences, paragraphs, or documents without the supervision of the annotated answers.

    It is important to emphasize that from Definition \ref{def.qas} follows that the meaning of Q and KS is not symmetric. Instead, the knowledge source provides a form of parametrization of the prediction model M. Hence, different questions $q_i$ can be asked without changing the knowledge source.

    Finally, in order to evaluate a predicted answer one needs a gold answer (also called a reference, labeled answer, or ground truth) corresponding to ${ r_i \in R}$. Here $R$ is the reference set containing all gold answers.

\begin{definition}[Error score of a QA system]
    An error score of a QA system is a function that maps the input space, $(A, R)$ or $(A)$ respectively to case 1 and case 2 of definition \ref{def.ks}, to the output space $S$, corresponding to error scores, by
    
\begin{eqnarray}
	E_{Case1}: (A, R) \rightarrow S
\end{eqnarray}

\begin{eqnarray}
    E_{Case2}: (A) \rightarrow S
\end{eqnarray}   

whereas a particular pair $(a_i, r_i) \in (A, R)$ is mapped to a particular score, $s_i \in S$, by

\begin{eqnarray}
s_i = E_{Case1}(a_i, r_i).	
\end{eqnarray}

\begin{eqnarray}
s_i = E_{Case2}(a_i).	
\end{eqnarray}
\end{definition} 

    The output space of scores $S$ can assume different forms. For instance, a score $s_i$ can either assume values between zero and one.
    
\begin{eqnarray}
S = [0,1],
\end{eqnarray} 

or it is either zero or one, i.e.,

\begin{eqnarray}
S = \{0,1\}.
\end{eqnarray} 
    An example of an error score of the former is the precision and for the latter the exact match (EM). Both error scores and many more will be discussed in detail in Section \ref{sec.errorscores}.

\section{Question Answering Paradigms}\label{sec.paradigms}

    It is clear that the number of different QA systems that can be established for different combinations of the main components of a QA framework described in Section \ref{sec.qaframework} is enormous. Interestingly, all of these realizations can be categorized into three QA paradigms that are dominating the design of current QA systems: (1) Information Retrieval-Based Question Answering, (2) Knowledge Base Question Answering, and (3) Generative Question Answering.

	\subsection{Information Retrieval-Based Question Answering }\label{sec.irqa}
	
    The first paradigm we discuss is Information Retrieval-based Question Answering (IRQA) which is also known as open-domain QA. It is widely used and the most typical paradigm for many commercial applications such as IBM's Watson \cite{45} or search engines like Google.  
	
\begin{figure}[!t]
		\centering
		\includegraphics[width=1\textwidth]{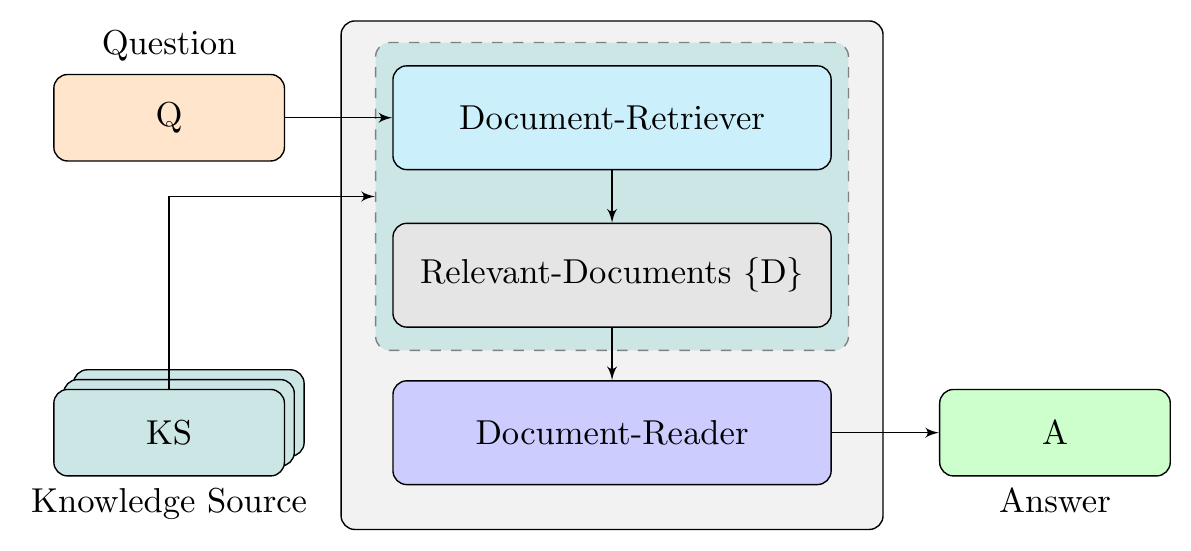}
		\caption{Architecture of information retrieval-based question answering (IRQA). In an IRQA the document-reader processes the query using different approaches(i.e.,  parsing, focus detection, relation extraction, lexical answer type detection, named entity tagging, question classification, etc). Furthermore, it retrieves the most relevant documents or passages from an available knowledge source (structured or unstructured data) and scores the candidate's answers. The document-reader is used for reading the retrieved documents and extracting the response.}
		\label{fig:GA}
	\end{figure}

    IRQA tries to predict the intent that may exist in the question, in order to identify the most relevant documents among available collections \cite{14}. It is not just taking a query in natural language form to find a list of relevant documents, but it provides an extra layer for handling those relevant documents and for extracting the answer by matching identified intent found in the question within the responses that may exist in the database. Usually, this paradigm includes two fundamental components called document-retriever and document-reader as depicted in Figure \ref{fig:GA}. In the following, we describe their functioning. 
	
\subsubsection{Document-retriever}

    The purpose of a document-retriever is to process the query and retrieve the most relevant documents, then rank every retrieved document to a given query (question) based on similarity score. This process starts with weighting the terms of each document using, e.g., Term Frequency-Inverse Document Frequency (TF-IDF) \cite{88,89} or Okapi Best Matching (BM25) \cite{57,87} or any other dense weighing mechanisms, to identify the informative terms among the most relevant documents. Then both query and retrieved document vectors need to be normalized, and a similarity score is computed between these vectors using e.g., the cosine similarity or the dot product. In order to efficiently improve the search speed and rank the most relevant documents that contain informative data, like the document frequency or term counts, an inverted index or alternative indexing technique based on e.g., hashing algorithm needs to be used\cite{47,56}. However,  the dictionary and postings list are the main components of the inverted index that hold the given query term frequencies and the list of document IDs associated with each term respectively.

\subsubsection{Document-reader}


    A document-reader is also called a {\it comprehension algorithms} that deals with different responsibilities, such as parsing the query {\it Q}, reading the relevant documents {\it D} that have already been retrieved by the document retriever from the available knowledge source (e.g., open-domain datasets), selecting the scored passages, and extracting the appropriate answer {\it A}. The document reader performs these tasks, which usually start with detecting the focus and extracting the relation from the question itself, and then, formulating certain keywords or embeddings in order to find a possible answer to the given question among the retrieved documents by computing a probability as \(P(a\vert{q},p)\). 
	
	Moreover, the positions of the start and end tokens of the answer could be represented as \(a_{s}\) and \(a_{e}\) respectively, and the probability of finding each token can be written as follows:
	
	\begin{equation}
		P(a\vert{q},p)= P_{s}(a_{s}\vert{q},p_{i}) \times P_{e}(a_{e}\vert{q},p_{i})
		\label{eq:pOD}
	\end{equation}

    Usually, the knowledge source (i.e., relevant document) needs to be retrieved by document-retriever, and then it should be passed along with the questions to QA comprehension algorithms, to train the document-reader component for predicting the candidate answer(s).  
    
	However, during the testing phase, the system is given only the question without the passage. The document-retriever component is expected to search for the relevant passages in the entire web resources or Wikipedia corpus and then feed these retrieved passages to the document-reader in order to produce the answer, either in an extractive form (e.g., a span of a text) or in an abstractive form.

\begin{figure}[h]
		\centering
		\includegraphics[width=1\textwidth]{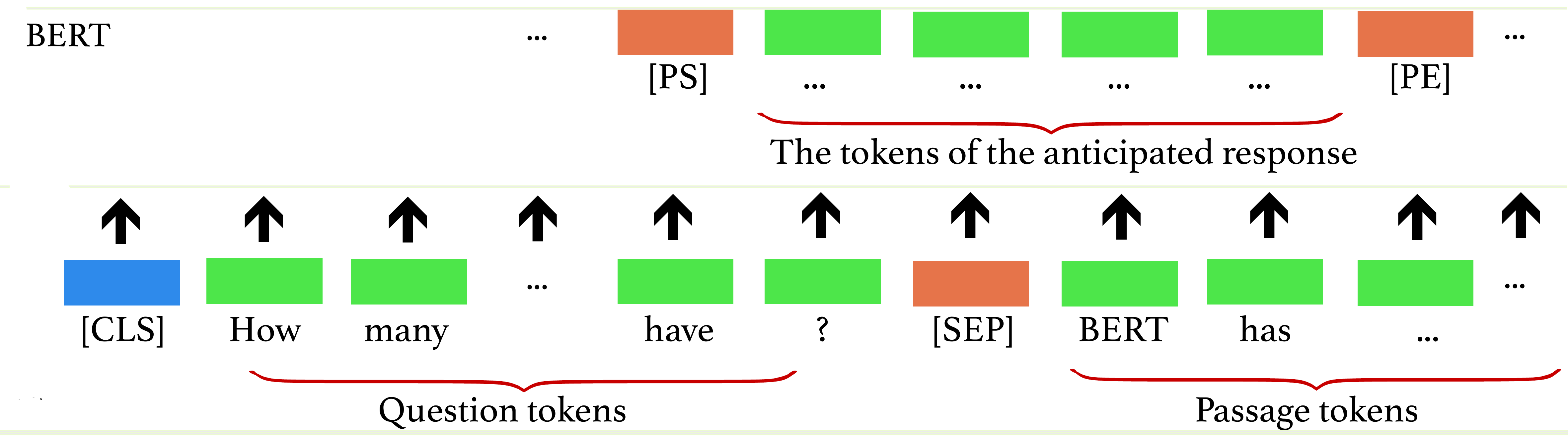}
		\caption{An illustration of a simple training phase of an IRQA as used by BERT. Here the associated question and the passage are treated as encoded strings separated by a [SEP] token, and the output word vector is encapsulating the sequence of tokens that are located between the start (PS) and the end position (PE) and represents the anticipated response.}
		\label{fig:BertOD}
	\end{figure}

	Alternatively, recent transformer-based algorithms treat the associated questions and passages as encoded strings, separated by a [SEP] token, in order to produce an embedding token for each passage token \(p_{i}\), as shown in Figure \ref{fig:BertOD}. 
	
	The positions of start and end tokens of the possible  {\it answer} in each passage \(p_{i}\), can be predicted by learning the start {\it S} and end {\it E} embedding vectors through a trained linear layer (e.g. SoftMax), during the fine-tuning phase as shown in Equations \ref{eq:b1} and \ref{eq:b2}:
	
	\begin{equation}
		P_{s}= \frac{exp(S.{\acute{p}_{i}})}{\sum _{j}exp(S.{\acute{p}}_{j})} 
		\label{eq:b1}
	\end{equation}
	
	\begin{equation}
		P_{e}= \frac{exp(E.{\acute{p}_{i}})}{\sum _{j}exp(E.{\acute{p}}_{j})} 
		\label{eq:b2}
	\end{equation}
	
	Here \(p_{s}\) and \(p_{s}\) are the probabilities of the span-start, and the span-end respectively, which could be estimated by finding the cosine similarity between the start or the end vector and each output token in the passage \(\acute{p}_{i}\) and then normalize the result over all the output tokens in the passage \(\acute{p}_{j}\).

	For estimating the training loss of each instance, the Log-Likelihood function (LL) is used to compute the negative sum of probabilities of the start \(P_{s}\) and the end \(P_{e}\) positions, as shown in equation \ref{eq:LL}: 
	\begin{equation}
		LL= -log P_{s} - log P_{e}.
		\label{eq:LL}
	\end{equation}

    A possible problem is that some datasets may not include the answer within their passages \cite{44,50}. In this case, the [CLS] token will be used as the answer with a start and end pointer for a given question \cite{58}. In addition, a concatenation for each observation may be required \cite{156} when the 512 input tokens of BERT are smaller than the annotated gold passages of some datasets \cite{50}. Hence, all observations in the entire document/passage will be scanned with a sliding window of size 512, to find the span answer, and the observation will be marked if an answer is found, otherwise, no label will be obtained. If more than one observation is found, the final answer will be selected from the observations with the highest probability. Some QA systems may use hybrid knowledge sources, see, e.g., \cite{45}. 
	
    Most modern IRQA systems utilize neural-based techniques, which require huge amounts of data. Unfortunately, the domain knowledge is not fully utilized, the available datasets still suffer from a limited size, and annotating process is very expensive.

    In order to enhance the capabilities of IRQA systems and to avoid the vocabulary mismatch problem in the extractive QA task \cite{59}, several techniques have been proposed. For example, instead of using the sparse vector of word counts for editing the distance or finding the exact match between the sequence of words in the query and answer, a dense embedding vector was introduced to find the semantic similarity between the candidate answer and query \cite{154,124}.

     in addition, a more sophisticated neural network architecture with two encoders has e been proposed to address the vocabulary mismatch issue by encoding the query and the context \cite{60,61,155}. Moreover, instead of using a [CLS] token for representing the encoded text, variant models of BERT have been used along with an average pooling layer over the output of all tokens, to add extra encoding weight and compute the similarity \cite{61}.

	Transformer models like BERT \cite{58}, RoBERTa \cite{146}, ALBERT \cite{147}, distilBERT \cite{148} have been employed as pre-trained models with a large number of example (training) questions and answers and have shown promising results. These models can also extract the answer of a given query from the collection of documents directly without the need to manually define features like the answer type. However, those models require massive computing resources along with huge training datasets. 
 
	Moreover, some of these transformers have adjusted their encoders to enhance the dense vector representations in order to deal with long documents and to improve IRQA components \cite{162,165}.
	
	Finally, Only IRQA paradigm has a document-retriever component, while the other paradigms like knowledge-based and generative question answering that will be discussed in the following sections, do not have it, and involve just the document-reader component.

	\subsection{Knowledge Base Question Answering }\label{sec.kbqa}

    Another main paradigm of QA systems is Knowledge Base Question Answering (KBQA) which is also known as closed-domain QA. A KBQA retrieves the answer directly from a specific knowledge-base; and eliminates the need for the document recovery component, which is an essential component of IRAQ.

    In order to accomplish this, KBQA relies on an internal search using parse or semantic vector representations to map the query to a local database \cite{33}. For realizing such a mapping there are two common approaches, the first uses a \emph{graph-based} model and the second \emph{semantic parsing} \cite{62,63,64,65}. Such algorithms can be designed to address task-specific requirements \cite{90,91,52,93,94}, e.g., by detecting frequently observed patterns or relationships. This can be done in a supervised or unsupervised manner.

    Supervised learning methods are used to process the query against the question-answering pairs dataset and create required segments of the parsing tree. Then it maps the relevant entities, e.g., parts-of-speech (nouns, verbs, and modifiers) to an appropriate form and matches the segments with existing relationships. 

    At this point, the KBQA model needs to gather more data from the parsing tree, and map it to more complex and coherent queries. It begins with identifying the subject of the question and linking it to the subject entity of the knowledge base. The answer is then derived by implementing a parsed logic model or via a graph representation. 

    KBQA based on closed-domain can be more coherent and interpretable by including dictionaries that are able to capture the semantics and the structure of a natural language. Recently, studies have begun to pay more attention to answering complex KB questions \cite{95,96} that comprise small scope entities or those with numerous subjects, complex relationships, multi-hops reasoning along with numerical operations, and compiled further the final answer by choosing the possible candidates \cite{128, 94,129}. To address such tasks, KBQA has used a variety of mechanisms, either by representing the overall information related to the question by means of a specific graph, e.g., a knowledge graph, that is capable of classifying all the entities extracted based on their relevance to a question, or by using a symbolic logic model to represent the question and then apply it to the knowledge-base to obtain the final answers \cite{93,52,91}.

\subsection{Generative Question Answering }\label{sec.gqa}
	
    In recent years, a third paradigm is becoming increasingly popular, which is called Generative Question Answering (GQA) \cite{171}. This paradigm seeks to directly generate a response according to the given input \cite{67,36}. The response can be generated as standalone, sequential or multiple choice question \cite{107,67,68}, based on different levels of context such as: keywords \cite{74,75}, sentences \cite{76,77,173}, paragraphs \cite{69,70}, documents \cite{80,81}, or multiple documents \cite{82,83}.

    In the literature, one can find many studies employing a large variety of architectures with different levels of complexity along with bulky features and auxiliary mechanisms in order to generate a suitable response from a given input (i.e., passages or documents). However, deep learning models with a sequence-to-sequence architecture and transformer models are becoming the most widely used technique \cite{9}. The sequence-to-sequence architecture has been built using different models and employs two layers of sequence neural network such as RNN or LSTM \cite{7,9}, one as an encoder for the input source context, and the other as a decoder for the embedded information (context vector) of the generated response \cite{67}. A long input sequence can cause a problem with RNN models; since RNN can forget such long sequences during the training phase due to the gradient vanishing phenomenon that hindered the model from updating its learning weights with too small gradients \cite{6}. Such architecture has been further extended to be adjusted with block cells to tackle the gradient vanishing problem \cite{172} and with more linguistic features \cite{68}. The attention-based model also has been introduced to investigate the effect of encoding sentences against paragraph-level information \cite{69,70}. Moreover, reinforcement models based on policy gradients have been proposed for QA to obtain good results \cite{71,166}. To mitigate the memory bottleneck of RNNs \cite{72}, a transformer over RNNs \cite{11,72,73,164} has been adopted to focus on a particular context in a paragraph. Despite the fact that most of the above-mentioned techniques were found to be more robust compared to the traditional techniques, they were also found to be more complex and hungry for more data (it demands huge amounts of data along with vast computational resources). For this reason, fine-tuned transformers with a pretrained language model \cite{73,165} were proposed to diminish some needs for data and computational resources.

	\section{Benchmark Data and Evaluation Scores} \label{sec.Benchmark}
	
    Since QA comprises many complex tasks with many variations, the benchmarking of QA is not straightforward. For this reason, task-specific datasets are needed in combination with expressive error scores. Therefore, drawing a conclusion about the quality of a QA system requires an informed assessment using labeled (annotated) data together with quantifiable scores that allow a comparison of the performance between different QA systems. In the following section, benchmark datasets and evaluation scores are discussed which are widely used in the literature.

	\subsection{Benchmark Datasets}\label{sec.benchdata}
	
	Many datasets have been proposed for benchmarking QA systems. Such datasets vary in their formats, types, the number of questions, and the related answers they contain. Each question of such datasets may be associated with one or more answers or none. When there is no answer associated with a question, the answer is usually marked by the annotators as '\emph{none}' as a gold answer to be returned during the evaluation process. Some QA systems based on neural techniques (e.g., end-to-end models) try to limit the number of duplicate questions in each subset of a dataset by randomly dividing the question-answer pairs into three subsets, known as training, development, and testing set, where the development set is used for fine-tuning the hyperparameters of a selected model, see \cite{69,169}.
	
	Usually, a QA system needs to be trained upon such datasets in order to be able to produce an answer or generate a desirable response. For this reason, several efforts have been made by researchers to create different types of datasets to be investigated with different QA paradigms. For example, Table \ref{tab:DB} shows a list of document-based datasets that can be used for IRQA, and Table \ref{tab:KB} shows another list of datasets that can be used for KBQA.

	\begin{table}[h!]
		\caption{ A list of document-based datasets that contain a set of passages and associated questions. The column 'Size' gives the number of questions and answers in the corresponding datasets and 'Description' provides information about the application domain. } 
		\centering 
		\begin{tabular}{p{2cm} p{4cm} p{4cm} p{1cm} } 
			\hline\hline 
			Dataset & Size & Description & Reference\\
			\hline
			BMKC-LS & 369780 questions & Biomedical dataset& \cite{21}\\
			cMedQA v2.0 & 180K questions, and 203569 answers& Medical dataset & \cite{18}\\
			MedQuAD & 47457questions, and 47457 answers& Medical dataset & \cite{19}\\
			StackExchange cQA & 20278 questions, and 82260 answers & Financial dataset & \cite{20}\\
			TOEFL & 963 questions, and 3852 answers & Listening Comprehension Test dataset & \cite{22}\\
			WebMedQA & 63284 questions, and 316420 answers& Medical dataset & \cite{112}\\
			\hline
		\end{tabular}
		\label{tab:DB}
	\end{table}

	\begin{table}[h!]
		\caption{A list of knowledge-based datasets that contain a set of entities and associated relations. The column 'Size' gives the number of entities and relations in the corresponding datasets and 'Description' provides information about the application domain.} 
		\centering 
		\begin{tabular}{p{2cm} p{4cm} p{4cm} p{1cm} } 
			\hline\hline 
			Dataset & Size & Description & Reference\\
			\hline
			LC-QuAD 2.0& 21258 entities, and 1310 relations& large scale complex question answering dataset &\cite{25}\\
			MetaQA&43233 entities & A movie dataset with multi-hops questions &\cite{26}\\
			PathQuestion&2215 entities, and 364 relations& A Freebase dataset with 2-hop path&\cite{30}\\
			PathQuestion Large&5035 entities, and 14 relations& A Freebase dataset with 3-hop path&\cite{30}\\
			SPADES&4K entities& A dataset that contains question types with blanks to fill in&\cite{28}\\
			WorldCup2014&1127 entities, and 6 relations& A football players knowledge base with multi-hops reasoning. &\cite{27}\\
			\hline
		\end{tabular}
		\label{tab:KB}
	\end{table}
	In general, knowledge base datasets are created by considering the question subject as the topic entity in a knowledge base, while various templates may be used to pre-create simple questions. Relying on these templates, more complex questions can also be created,  as well as by using an executable logic template. In addition, a set of rules can also be used to extract the response. Besides, to further improve the generated logical models and generate natural language questions that appear more diverse and fluent, crowd workers are hired to reformulate the queries.
	
    Creating a high-quality data set dedicated to generative question answering is an important direction that can significantly enhance the development of QA systems. Popular GQA datasets can be classified based on the type of target answers and generated questions, as shown in Table \ref{tab:SaSim} (standalone questions with simple answers), Table \ref{tab:SaAbs} (standalone questions with abstractive answers), Table \ref{tab:SaAgo} (standalone questions with agnostic answers), Table \ref{tab:SeAbs} (sequential questions with abstractive answers), Table \ref{tab:MuSim} (multiple choice questions with simple answers) and Table \ref{tab:MuAbs} (multiple choice questions with abstractive answers.). Moreover, many of these datasets have been also tailored to be used for information retrieval-based or knowledge-base QA systems.

	\begin{table}[h!]
		\caption{GQA datasets for standalone questions with simple answers. The column 'Size' gives the number of examples (i.e., question-answer pairs) in the corresponding datasets, and 'Description' provides more information about the datasets.} 
		\centering 
		
		\begin{tabular}{p{2cm} p{4cm} p{4cm} p{1cm} } 
			\hline\hline 
			Dataset & Size & Description & Reference\\
			\hline
			SQuAD 1.1 & 107,785 question-answer pairs &   Collected as crowdsourced QA pairs from Wikipedia from 536 Wikipedia articles. & \cite{17}\\
			NewsQA& 120,000 question answer pairs & Collected as crowdsourced QA pairs from 10K CNN news articles. & \cite{97}\\
			SearchQA& 140,000 question answer pairs & Each pair has 49.6 snippets on average. & \cite{98}\\
			HotpotQA& 113,000 question answer pairs & Wikipedia-based QA pairs with questions and answer reasoning.& \cite{48}\\
			Natural Questions & 307,373 training examples, 7,830 development examples, and 7,842 test examples. &Each example consists of a Google query (i.e., question) and a related answer within the Wikipedia page.& \cite{50}\\
			TriviaQA& 650K question-answer-evidence triples.& Collected as triples of question-answer pairs from different Wikipedia and Web pages.& \cite{49}\\
			\hline
		\end{tabular}
		\label{tab:SaSim}
	\end{table}

	\begin{table}[h!]
		\caption{GQA datasets for standalone questions with abstractive answers. The column 'Size' gives the number of examples in the corresponding datasets and 'Description' provides information about the datasets.} 
		\centering 
		\begin{tabular}{p{2cm} p{4cm} p{4cm} p{1cm} } 
			\hline\hline 
			Dataset & Size & Description & Reference\\
			\hline
			MS MARCO & 1010916 questions, 8,841,823 paragraphs retrieved via Bing from 3,563,535 web documents.
			& The answer for each question has been manipulated by crowdsourced, and around 182669 answers were completely rewritten. & \cite{102}\\
			NarrativeQA &46765 QA pairs written by crowdsourced. & The script was gathered as 1567 stories from different resources i.e., books and movies. & \cite{103}\\
			DuReader& 200K questions, along with 420K answers. & An open-domain dataset 
			with 1M documents where answers are manually generated.
			& \cite{104}\\
			SQuAD 2.0 & 150K& This dataset combines the 100,000 questions in SQuAD1.1 with over 50,000 un-answerable questions. & \cite{44}\\
			\hline
		\end{tabular}
		\label{tab:SaAbs}
	\end{table}

	\begin{table}[h!]
		\caption{GQA datasets for standalone questions with agnostic answers. The column 'Size' gives the number of examples in the corresponding dataset and 'Description' provides information about the dataset.} 
		\centering 
		\begin{tabular}{p{2cm} p{4cm} p{4cm} p{1cm} } 
			\hline\hline 
			Dataset & Size &Description & Reference\\
			\hline
			LearningQ & 230K document-question pairs. & Based on popular online learning platforms. & \cite{106}\\
			\hline
		\end{tabular}
		\label{tab:SaAgo}
	\end{table}

	\begin{table}[h!]
		\caption{GQA dataset for sequential questions with abstractive answers. The column 'Size' gives the number of examples in the corresponding dataset and 'Description' provides information about the dataset.} 
		\centering 
		\begin{tabular}{p{2cm} p{4cm} p{4cm} p{1cm} } 
			\hline\hline 
			Dataset &Size & Description & Reference\\
			\hline
			CoQA & 127K conversational questions with answers.
			& Designed to enable machines to respond to a series of questions that appear in conversation, based on 8K conversations about texts gathered from seven different domains.
			& \cite{107}\\
			QuAC &14K information-seeking QA dialogues (100K questions in total). & In order to make QuAC more like search engine queries that are retrieved from users’ dialogues, this dataset consists of large-scale dialogues of Wikipedia articles, in which crowd-workers do not know the answer to their questions ahead of time.
			& \cite{108}\\
			\hline
		\end{tabular}
		\label{tab:SeAbs}
	\end{table}

	\begin{table}[h!]
		\caption{GQA dataset for multiple choice questions with simple answers. The column 'Size' gives the number of examples in the corresponding dataset and 'Description' provides information about the dataset.} 
		\centering 
		\begin{tabular}{p{2cm} p{4cm} p{4cm} p{1cm} } 
			\hline\hline 
			Dataset &Size& Description & Reference\\
			\hline
			CBT& 687343 questions
			collected from 108 books.
			& It aims to measure the robustness of language models based on a wider linguistic context.
			Each question is provided with up to 10 possible answers that appear in the context sentences as well as the query. & \cite{109}\\
			\hline
		\end{tabular}
		\label{tab:MuSim}
	\end{table}

	\begin{table}[h!]
		\caption{GQA datasets for multiple choice questions with abstractive answers. The column 'Size' gives the number of examples in the corresponding datasets and 'Description' provides information about the datasets.} 
		\centering 
		\begin{tabular}{p{2cm} p{4cm} p{4cm} p{1cm} } 
			\hline\hline 
			Dataset & Size\textbf{Number of Questions }& Description & Reference\\
			\hline
			RACE& 97687 multiple choice questions along with 27933 passages.
			& Large-scale reading comprehension dataset collected from English exams for middle and high Chinese schools consists of nearly 28,000 passages and nearly 100,000 questions generated by human experts, where each question has four possible answers, and only one answer is correct.
			& \cite{110}\\
			MCTest & 500 fictional stories and 2000 questions. & a freely available set of stories and associated questions. Each story has four multiple-choice questions. The stories and questions are selectively chosen to reflect what a young child would understand. & \cite{111}\\
			\hline
		\end{tabular}
		\label{tab:MuAbs}
	\end{table}

	\subsection{Taxonomy of Evaluation Scores}\label{sec.errorscores}
	
	Every natural language is rich in words and synonyms, which makes it possible to produce a sheer unlimited combination of output sentences that can include different phrases with vastly different interpretations. Hence, due to this diversity of language expressions, the generated output of a QA system may involve different forms of ambiguities that can exist in many forms, e.g., {\it lexical, semantic, syntactic, etc.}; The generated output could be less informative and difficult to understand due to these ambiguities, which may add some complexity and vagueness in terms of {\it fluency, correctness, and coherence} to the language. However, the goal of NLG is to obtain a more valuable and coherent output text that has more readability and understandability. Thus, the system's output ({\it i.e.,  hypothesis, generated/predicted response/answer}) needs to be evaluated against the given reference (also called {\it ground truth, gold answer, labeled response}) in order to determine its quality \cite{137, 5}. However, due to the complexity of natural language, such output is a structured object rather than a single number, which makes the evaluation itself a complex task.

\begin{figure}[t!]
		\centering
		\includegraphics[width=0.95\textwidth]{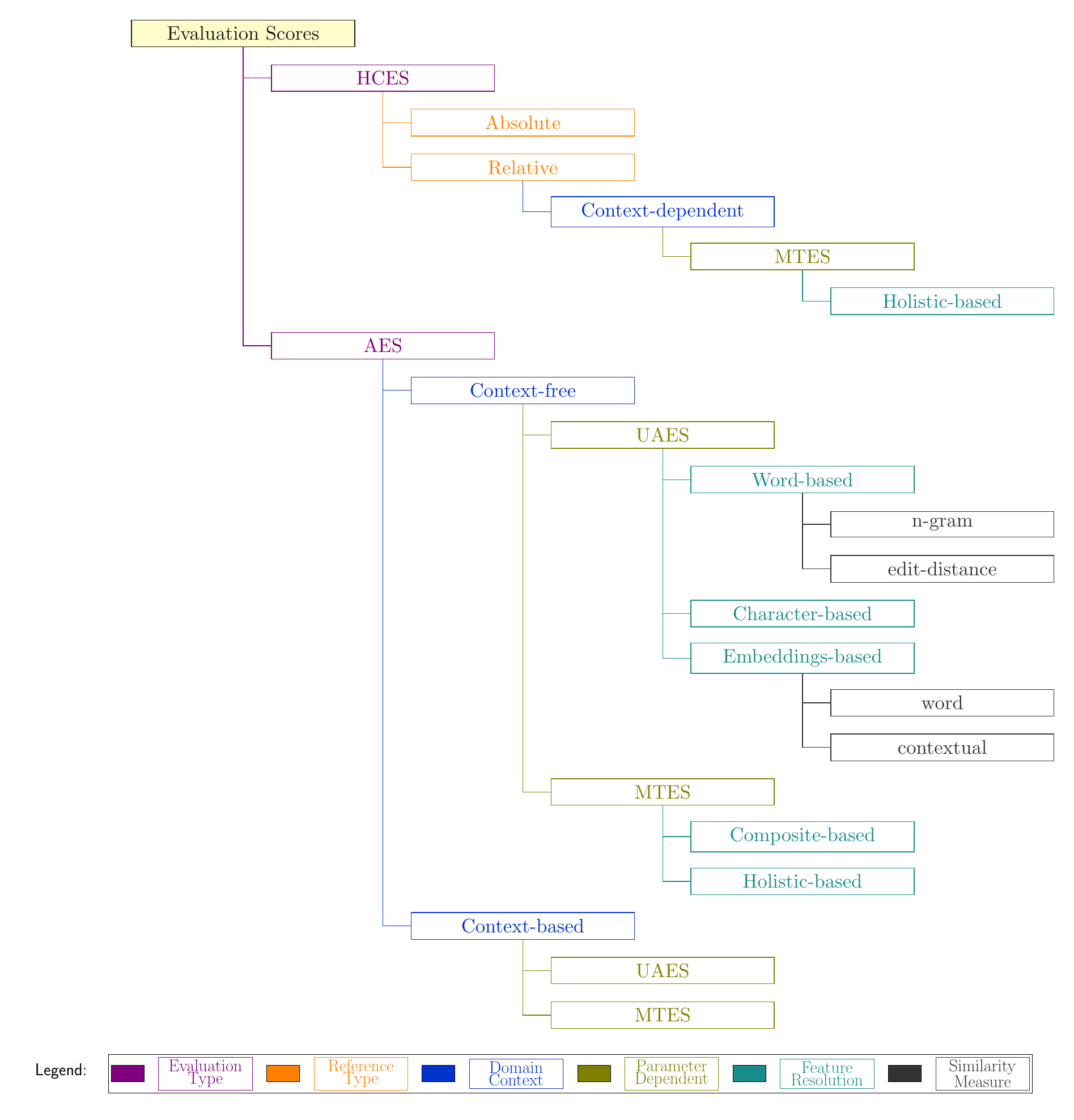}

		\caption{The taxonomy of evaluation scores. The used acronyms have the following meaning: HCES: Human-Centric Evaluation Scores,  AES: Automatic Evaluation Scores, UAES: Untrained Automatic Evaluation Scores,  MTES: Machine-Trained Evaluation Scores. The legend indicates the type of score in the branches.}
		\label{fig:EM}
	\end{figure}

	For this reason, many different types of evaluation scores have been proposed \cite{5, 136,139, 140}. Overall, these scores can be broadly classified into two categories, see Figure \ref{fig:EM}:
	
	\begin{itemize}
	 \item[] (i) {\bf Human Centric Evaluation Score (HCES)}
	 \item[] (ii)  {\bf Automatic Evaluation Score (AES)}
	\end{itemize}

	Every category has its own strengths and weaknesses. In general, a {\bf HCES} is considered the best evaluation that gives the most trustworthy score. Typically, HCES relies on a group of people (e.g. experts or specialists) to evaluate the output of a QA system based on certain guidelines and criteria, e.g., regarding adequacy, fluency, and coherence of a text. However, it is also prone to human errors caused by subjectivity and bias in opinion. Frequently, a 5-point Likert-scale \cite{199} or numerical ranking criteria is used as a measurement tool for HCES evaluations \cite{202}.  Conceptually, HCES can be further classified into two sub-categories: {\it Absolute} evaluation and {\it relative} evaluation.
	
	{\bf Absolute evaluation:} In this type of HCES, individuals are asked to evaluate the output of a natural language generation/inference model using a 5-point scale \cite{203}. The evaluator assesses the output texts of a task based on predefined criteria or rubric, rating each from 1-5 or on a nominal scale, For example, a statement such as "Rate the output based on the naturalness of the language" with options as \textit{very unnatural, unnatural, neutral, natural and very natural}, or with 1-5 numerical options, each responding to a similar semantic meaning. The results of which will be averaged over the rubric and used as the evaluation/annotation output. For example, in machine translation, the task may require an evaluation from a person who is familiar with either the target language (i.e., monolingual) or both source and target languages (i.e., bilingual) to perform the required assessment (rating) on the generated text. The absolute method is often subjected to the bias of the evaluator due to the discreteness of the scale and inter-annotator disagreements. Therefore, individual evaluators may not always be preferred for some tasks, e.g., in order to obtain a more accurate assessment.

	{\bf Relative evaluation:} In this approach, the task output is not given an exact absolute score, instead given several outputs to a given task, the evaluator is asked to order them based on the quality of the output \cite{202}. often tasks with lengthier outputs (e.g., dialogue generation, machine translation) requires that they be evaluated as a whole conversation or a whole translation. In this case, either experts or crowd-source workers could be asked to rate the output of a system (i.e., the multiple responses or comparing the predicted response with the multiple references), by ranking each output as compared to each other (for example $n$ different conversations with a chatbot from most human-like to least human-like). Relative evaluation is preferred when the speed of assessment matters. However, crowdsource workers are not preferred when a given task does require domain knowledge (e.g., in a medical field). 
	
	The scores acquired by each evaluator can be computed as a weighted average based on their past performance, or aggregated as a simple average \cite{raykar2010learning}. In order to ensure the reliability of the evaluation or/and annotations, HCES are subjected to reliability measures such as intra-observer agreement (IA), and inter-annotator agreement (IAA). In the case there are clearly defined expected evaluation outputs, the accuracy of the evaluation can also be a measure of data reliability. In general, a high inter-annotation agreement is desirable, which measures the agreement between several evaluators on the quality ranking of an output. Whereas, the intra-observer agreement is used as a reliability measure when the evaluation or annotation is done by a smaller group of experts over a long period of time. Here a certain amount of time is let to pass after the initial evaluation, thereafter same evaluator is asked to rate the same outputs. If the evaluator's previous and current ratings have a high agreement then the data is considered reliable. The IAA is the most utilized data reliability measure in HCES due to convenience and in literature, several a coefficient of the agreement have been introduced to measure IAA such as Cohen's Kappa, Fleiss Kappa, or Krippendorff's alpha \cite{198}. However, achieving a sufficiently high IAA is difficult due to human error, insufficient guidance or preparation, and ambiguity in the generated response \cite{5, 143}.

	Despite its obvious benefits and superiority. HCES is expensive, time-consuming, tedious, and requires domain expertise, which makes it an impractical choice when dealing with a task with a larger data output or some dynamic behavior. However combining HCES with MTES, where the trained evaluator models receive some feedback from human annotators, a more practical success can be achieved. An example of this is ADEM \cite{150}, which will be discussed later in the section.

	Unlike HCES, the {\bf AES}  are cheaper evaluation techniques (e.g., do not require domain experts as the HCES do) and found to be more practical when dealing with a large task that may involve some dynamic behavior  \cite{144}. Importantly, such scores are usually correlated well with human judgments \cite {137}.

	AES can be further sub-categorized into: {\it context-based} and {\it context-free} evaluation, discussed below, depending on the utilized mechanisms of the evaluation process.

	A {\bf context-free} evaluation does not consider the context during the evaluation process. Instead, it just considers the sequences of words (e.g., n-grams) among the obtained output and the reference in order to assess the overlap or match. For this reason, this evaluation category can be seen as tasks-agnostic and easily adopted to a wider range of QA systems. In contrast, a {\bf context-based} evaluation is specifically designed to address a particular requirement of a specific problem and cannot be easily adopted for different problems like the {\it context-free} evaluation.

	Both {\it context-free} and {\it context-based} AES scores can be further sub-categorized into: 
	\begin{itemize}
	\item[] (i) {\bf Untrained Automatic Evaluation Score (UAES)}
	\item[] (ii) {\bf  Machine-Trained Evaluation Score (MTES)}
	\end{itemize}

	A {\bf UAES} does not require any pre-training because such scores do not depend on pre-defined parameters. The lack of parameters makes UAES easy to use, so it becomes very popular. Such category can be further sub-categorized based on the operation level among the text units, which can be:
	
	\begin{itemize}
	\item Character-based
	\item Word-based
	\item Embeddings-based
	\end{itemize}

	A {\bf character-based} score shows promising performance in evaluation task for machine translation \cite{177}. Importantly, this error score is simple enough to directly work on the response and reference strings, and does not require tokenization.

	A {\bf word-based} score treats the response and the reference as a {\it bag-of-words} and does not usually consider the context of an input text for the evaluation. Instead, it compares and evaluates the response against the available reference via a set of heuristic features,  such as {\it Precision} or {\it Recall}, and computes the similarity between text units based on the matching overlap, or based on the number of word edits required to make the response similar to the given reference \cite{174,99}. The  {\it word-based} scores include a wide range of error scores, such as BLEU \cite{134}, METEOR \cite{135}, ROUGE \cite{141}, NIST \cite{168} and CIDEr \cite{145}. 

	An {\bf embedding-based} score is not designed to simply rely on overlap matching between response and reference, as most of the word-based scores do, but to capture the semantic similarity (i.e., word or distributional similarities) that may exist between answer and reference \cite{175,176,178, 142}.

    In contrast to UAES, a {\bf MTES} contains adjustable components, i.e., parameters, which are specifically estimated for a given task via a learnable model. That means an MTES is a parametric model which needs to be trained based on data. MTES can be classified into two sub-categories: 
		\begin{itemize}
		\item[] (i) {\bf Composite measures }
		\item[] (ii) {\bf Holistic measures }
		\end{itemize}
	
    In the literature, composite measures are also called feature-based measures and holistic measures are called end-to-end measures. The difference is that a holistic measure cannot be separated into individual error scores while this is the case for feature-based measures where the features correspond to such individual error scores.

    {\bf Composite measures}: This category of error scores combines features using a parametric model whereas the features can correspond to any score in the categories S-UAES or A-UAES, e.g, precision, recall, or BLEU. Typically, a large number of such heuristic scores are used. For combining those features either simple parametric models are used, e.g., linear regression (see BEER \cite{179} or BLEND \cite{180}), or complex models based on neural networks (see \cite{181}).

    {\bf Holistic measures}: This category of error scores utilizes end-to-end models based on neural networks, e.g., transformers, to perform the evaluation. The end-to-end models can be directly applied to statistical or contextualized features that exist within the response or reference to compute the scores.  Such models may use different strategies to perform the evaluation process: (i) referenced scores, which compare the generated response with a provided gold answer, by learning an alignment score between context and response to approximate human judgments \cite{150}, (ii) hybrid referenced-unreferenced evaluation score \cite{149}, where the score is trained without needs for human responses by smoothing the negative samples directly from the dataset,  or (iii) unreferenced scores (i.e., learned in an unsupervised manner) that uses large pre-trained language models to extract latent representations of text units (i.e., utterances or words), and take advantage of the time shifts that exist between them \cite{153}. The MTES based on  {\it end-to-end}  models include a wide range of error scores such as BertScore\cite{142}, RUSE \cite{182}, RUBER \cite{149}, BLEURT \cite{151}, ADEM \cite{150}, MaUde \cite{153}.

\section{Quantitative error scores}\label{sec.ces}

    In the previous section, we introduced a taxonomy of evaluation scores providing an organization of the many score categories. In this section, we discuss specific scores thereof that are used as UAES and MTES. For reasons of clarity we further sub-categorize UAES into Simple Untrained Automatic Evaluation Scores (S-UAES) and Advanced Untrained Automatic Evaluation Scores (A-UAES).

	
    For the following presentation of the scores, we assume the predicted answer $A$ and gold answer (reference) $R$ are given by:
    
\begin{eqnarray}
    \text{predicted answer: } A &=& a_1, a_2,\dots a_m \label{eqn.A} \\
    \text{gold answer: } R &=& r_1,r_2,\dots r_n \label{eqn.R}
\end{eqnarray}

    That means $A =a_1, a_2,\dots a_m$ is a response (sentence) consisting of $m$ words $a_i$ and correspondingly $R=r_1,r_2,\dots r_n$ is the gold answer (sentence) consisting of $n$ words $r_j$. Depending on the context $a_i$ and $r_j$ can be words, unigrams, bigrams, or n-grams. Hence, our notation should be flexible enough to accommodate the common cases encountered. Hence, we assume A and R are tokenized and indexed according to Equation \ref{eqn.A} and \ref{eqn.R}.

\subsection{Simple Untrained Automatic Evaluation Scores (S-UAES)}\label{sec: S-UAES}
	
    In the category Simple Untrained Automatic Evaluation Scores (S-UAES), the most widely used error scores in the literature are an exact match (EM), precision, recall, F1-score, error rate (ER), average precision (AP), mean average precision (MAP) and mean reciprocal rank (MRR). In the following, we provide a discussion for each score.
	
	\begin{itemize}
		\item  {\bf Exact Match (EM)}: EM is a score that gives a one if the predicted answer is identical to the gold answers and a zero otherwise. That means EM is a binary score assuming values in $\{0, 1\}$.
		
		 \begin{example} 
			Suppose we have a labeled answer {R}:  "The quick brown fox jumps over the lazy dog",  and a predicted response  {A}  "the dog has slept on the mat".
			\label{example:1}
		\end{example}
		Based on the given definition of EM, the EM score of Example \ref{example:1} is $0$, since there are some differences between {A} and {R}. \\

 \item {\bf Precision}: The precision is the proportion of correct answers over all anticipated answers. It indicates the relevant returned documents.  

	\begin{equation}\label{key}
		Precision =\frac{\{relevant\ documents\} {\displaystyle \cap }\{retrieved\ documents\}}{\{retrieved\ document\}} 
	\end{equation}

Alternatively, the precision can be calculated as follows:
 
 \begin{equation}\label{key}
 	Precision =\frac{TP}{TP+FP} 
 \end{equation}
 
    where TP is the number of true positives and FP is the number of false positives. At that point, the precision can be computed as the number of n-grams that match between {A} and {R} divided by the total number of n-grams in {A}. On this basis, the computed precision of  Example \ref{example:1} is $3/7=0.43$, where  $3$ is the number of n-grams found to be matched between {A} and {R}, and  $7$ is the total number of n-grams of {A}. The n-gram match for precision is position independent.\\

	\item 	{\bf Recall}: The recall is the proportion of correct anticipated answers over the gold answer. It measures the relevant (correct) documents over all the actual possible documents. The output of such a score for each query is a binary value that indicates either or not the document is contained in the selection.

	\begin{equation}\label{key}
		Recall =\frac{\{relevant\ documents\} {\displaystyle \cap }\{retrieved\ documents\}}{\{relevant\ document\}} 
	\end{equation}
 
    Alternatively, the recall can be calculated as follows:

	\begin{equation}\label{key}
		Recall =\frac{TP}{TP+TN} 
	\end{equation}
	where TP is the number of true positives and TN is the number of true negatives. Therefore, the recall score could be computed as the number of n-grams that match between {A} and {R} divided by the total number of n-grams in {R}. Thus, the computed recall score of example \ref{example:1} is ($3/9=0.33$), where  $3$ is the number of n-grams found to be matched between {A} and {R}, and  $9$ is the total number of n-gram that {R} composed from.\\
	
	\item 	{\bf F1-score}: The F1-score aims to find the harmonic mean (balanced in importance)  between recall and precision. It measures the word overlap between the predicted and the gold answer in a more flexible way than that used in EM, as it allows us also to trade off precision against the recall. However, it does not capture the similarity between two correct answers that differ in their semantic content. The F1-score is defined as follows:

	\begin{equation}
		F1 = 2 * \frac{Precision * Recall}{Precision + Recall}
		\label{eq:f1}
	\end{equation}

	The F1 score  could be computed based on the equation \ref{eq:f1} as \(2 \times \frac {0.43 \times 0.33}{ 0.43 + 0.33  } =0.37\), where  $0.43$ and  $0.33$ are the score values of precision and recall respectively, that found over {A} and {R} in the given example.\\

	\item 	{\bf Error Rate (ER)}: The ER is the frequency or the probability of errors occurring when comparing A with R.
	
	\begin{equation}\label{key}
		Error \; Rate =\frac{FP+FN}{N} 
	\end{equation}
	where FP is the number of false positives, FN is the number of false negatives, and N is the total number of data points examined overall.\\

	\item 	{\bf Average Precision (AP)}: The AP is calculated as the weighted mean of precision for a query in the query set.
	
	\begin{equation}\label{key}
		AP = \frac{\sum _{x=1} ^{n} P(k)* R(k)}{number\ of\ the\ relevant\ documents}
	\end{equation}
	
	where {\it k} is the rank in the sequence of retrieved documents, {\it n} is the number of retrieved documents, {\it P(k)} is the precision at cut-off {\it k} ranked list documents, and {\it R(k)} is an indicator function equaling 1 if the item at rank {\it k} is a relevant document, zero otherwise.  The average is over all relevant documents and when the relevant documents are not retrieved it holds a precision score of zero. \\

	\item 	{\bf Mean Average Precision (MAP)}: The MAP is a score for evaluating the ranked retrieval list of answers, as the mean of AP scores for encountered relevant documents for each query.
	
	\begin{equation}\label{key}
		MAP =\frac{1}{N} \sum _{x=1} ^{n} AP_{i}
	\end{equation}
	
	where the {\it N} is the number of queries, and \(AP_{i}\) is the average precision score for encountered relevant documents for each query.\\
	
	\item 	{\bf Mean Reciprocal Rank (MRR)}: The MRR takes the position (rank) of an answer into account, evaluates the probability of correctness of each possible answer related to a sample of queries, and ranks them according to probability, where the inverted order of the query answer is the multiplicative inverse of the order of the first correct answer: {\it 1} for first place, {\it 1⁄2} for second place, and so on. Thus, the mean reciprocal order is the mean of the reciprocal orders of results of a sample of {\it Q} queries.
	
	\begin{equation}\label{key}
		MRR =\frac{1}{\vert{Q}\vert} \sum _{x=1} ^{\vert{Q}\vert} \frac{1}{rank_{i}}
	\end{equation}
	
	where \(rank_{i}\) is the rank position of the first relevant document related to the \(i^{th}\) query, while the reciprocal value of the mean reciprocal rank corresponds to the harmonic mean of the ranks.
	
	\end{itemize}

	\subsection{Advanced Untrained Automatic Evaluation Scores (A-UAES)} \label{sec:A-UAES}

	The next category of UAES is advanced scores. Among the most widely used context-free error scores are BLEU, NIST, ROUGE, and METEOR discussed in the following. All of these scores utilize precision, recall, and F1 score in some form.

\subsubsection{Bilingual Evaluation Understudy (BLEU)}

    BLEU compares the n-grams of the answer with the n-grams of the reference by counting the number of matches. These matches are position-independent. Obtaining the BLEUE score is a three steps process.
			
    First, the logarithmic geometric mean of the modified n-gram precision, $P_n$, is calculated using n-grams up to length $N$ and positive weights $w_n =1/N$ that summing to one

\begin{eqnarray}
    \log (GM) =  \Big( \sum_{n=1}^N \frac{1}{N} \log P_n \Big) = \log \Big( \prod_{n=1}^N  \log P_n \Big)^{\frac{1}{N}}
\end{eqnarray}

where $\Big( \prod_{n=1}^N  \log P_n \Big)^{\frac{1}{N}} = \prod_{n=1}^N  \log (P_n)^{w_n} $ is the geometric mean of the modified n-gram precision. Here the n-gram precision, $P_n$, is defined by

\begin{equation}
    P_{n}=\frac{\sum _{c \in C} \sum _{n-gram \in c} Count_{clip}( n-gram)}
    {\sum_{c^{'} \in C} \sum _{n-gram ^ {'} \in c^{'}} Count( n-gram^{'})} \label{mod.precision}
\end{equation}

    where {c} is the generated response (i.e., answer) that contains n-gram of text and appear in {C}, {C} is  set of the generated output (i.e., responses),  \(Count_{clip}(n-gram) =max(n-gram  \ \exists \ A) \subseteq R\), is the maximum number of times the given n-gram of {A} appears in any corresponding {R} references (i.e., gold answer), and  \(Count(n-gram)\)  is the total number of words in  {R}  and  \(\forall \ n-gram \in c^{'} \).

Second, a brevity penalty is calculated by
\begin{equation}
BP  =\left\{
\begin{array}{ll}
1, & \text{if } c > r \\
e^{(1-\frac{r}{c})}, & otherwise
\end{array}
\right.
\end{equation}
    where $c$ is the length of the predicted answer and $r$ is the effective reference answer length.

Finally, the BLEU score is calculated by
\begin{eqnarray}
BLEU = BP \cdot GM
\label{eq:bleu}
\end{eqnarray}

    As default parameters the original BLUE \cite{134} used $N = 4$ and uniform weights $w_n = 1/N$.


    In order to calculate the precision, BLEU starts counting up to the maximum number of times a word occurs in any single reference, it clips the total count of each candidate word, then, the modified n-gram Precision is equal to the clipped count divided by the total number of words in the candidate response.

\begin{example}
    Suppose we have two reference answers: (R1) "The dog sat on the mat", and (R2): "There is a dog on the mat". The predicted answer {A} is "the the the the the the the".
\label{example:2}
\end{example}

    In this case, all words of {A} appear in {R1} and R2, so the (ordinary) precision is $\frac {c}{c^{'}}$, where {c} is the number of words from {A} found in R1 or R2, and \(c^{'}\) is the total number of words in the generated answer A. As a result the (ordinary) precision of Example \ref{example:2} is $7/7=1$ for R1 and $7/7=1$ for R2.

    However, BLEU makes slight changes to the ordinary precision, as it takes into consideration the maximum total count of \(Count_{clip}(\text{\it n-gram})\) in any candidate references. Specifically, in Example \ref{example:2}, the word 'the' has a max count of $2$ as it appears twice in {R1}, hence, the clipped unigram count is $2$, and the total number of counts in R1 is $7$, thus, the modified n-gram precision according to Eqn. \ref{mod.precision} is $2/7$. For R2 a similar calculation gives $1/7$.

\begin{example}
    Suppose {R}  is "the fox" and {A} is "the the fox".
\label{example:3}
\end{example}

    Here, if the n-gram length  in example \ref{example:3} is $1$ (i.e., unigram), then the precision score will be ($1+1+1/3 = 1$), and if the n-gram  length is $2$ (e.g., bigram), then the precision score will be ($0+1/2 = 1/2$).
    Since the precision score depends only on the length on {A} and not of {R}, the evaluation system in this case can exploit the score and acquire high scores by obtaining only a few matching of n-gram with {R}. In other words, the score of evaluation can be high even if {A} is not the perfect match with any {R}. For this, the BLEU tries to prevent the short n-gram length of {A}  from receiving too high scores, by using the geometric mean to combine the n-grams scores and multiply it by a brevity penalty as shown in equation \ref{eq:bleu}.

\subsubsection{National Institute of Standards and Technology (NIST)}

    NIST \cite{168} is a modified version of BLEU originally proposed for evaluating machine translation. Unlike BLEU, which gives equal weight to each n-gram precision, NIST weighs matched n-grams differently based on an information gain to show how informative a particular n-gram is:
		\begin{equation}\label{key}
			\text{info}(w_{1},\dots, w_{n})= \log_{2} \frac{  \text{\it \# occurrences of } w_{1},\dots, w_{n-1}} {\text{\it \# occurrences of } w_{1},\dots, w_{n}}
		\end{equation}
    The evaluation score could be improved by assigning more weight to the rarely matched n-gram and lower weight to the common n-grams. For example, the bigram receives a lower weight than the rare n-gram "the brevity penalty". NIST uses an alternative brevity penalty as follows:

\begin{equation}\label{key}
			BLEU-N= \exp({\sum _{n=1}^{N} weight_{n} \log(Precision_{n})})  \cdot \exp(min(1-\frac{\vert r \vert}{\vert p \vert},0))
			\end{equation}
The final  NIST score is given by:
\begin{multline}\label{key}
NIST= \sum_{n=1}^{N} (\frac{ \sum _{\text{\it all matched n-gram}} {\text{\it \text{info}(n-gram)}}} { \sum _{ n-gram \in  A}}) \\
			\cdot \exp(\beta \log^{2} [min(\frac{\vert p \vert}{\vert r^{'} \vert},1)])
		\end{multline}

    Here, \(r^{'}\) is the average number of n-gram in  {R}, that averaged over all references, {p} is the number of n-gram in {A} being scored, and \(\beta \) is the chosen value to make BP factor =0.5, when the number of n-gram in {A}  is \(2/3^{rd}\) the average number of n-gram in {R}.

\subsubsection{Recall-Oriented Understudy for Gisting Evaluation  (ROUGE)}

    ROUGE \cite{141} is another modified version of BLEU that is based on recall instead of precision-based.

    There are four different versions of the ROUGE measure (ROUGE-N, ROUGE-L, ROUGE-W, and ROUGE-S) which are meant to automatically determine the quality of a response $A$ against the references $R$ in the abstractive summarization evaluation task. To evaluate the quality of a summarized text, those measures try to find how many n-grams in the referenced text (R) appear in the generated response (A).

    {\bf ROUGE-N:} The ROUGE-N score is an n-gram recall-based score that counts the n-gram matches between {A} and {R} by:

\begin{equation}\label{eq:rouge-n}
    \text{ROUGE-N} = \frac{\sum _{S \in R} \sum _{gram_n \in S} Count_{match}(gram_n)}  {\sum_{S^{'} \in R} \sum _{gram_n^{'} \in S^{'}} Count( gram_n)}
\end{equation}

    where {S} is the reference (i.e., ground truth)  which contains n-gram of text that appear in {R}, {R} is the set of the given references,  \( Count_{match}(gram_n)\) is  the maximum number of times the n-grams in {R} appear in any generated output {A} (i.e., response), \(S^{'}\) is the total number of words in the reference {R} and $n$ is the length of the n-grams. That means with ROUGE-N, the $N$  corresponds to the maximal n-gram length between the predicted answer and the given reference. 

    Returning to Example \ref{example:2}, we see that the word "the" is co-occurring twice in {A} and {R1} and once between {A} and {R2}. Thus, the ROUGE-N score according to Equation \ref{eq:rouge-n} can be calculated as the ratio of the number of unigrams in {A} found in {R}, over the total number of unigrams in {R}. This gives the ROUGE-1 scores for {R1} and {R2} of $2/6$ and $1/7$ respectively.

    Let's look at another example and calculate this time ROUGE-2 considering bigrams (i.e., n-gram=2).
    
\begin{example}
    Suppose that the given (R) is "The dog sat on the mat",  and the predicted response  {A}  was 'the dog and the cat'.
\label{example:4}
\end{example}

    In this example, we see that there is only one bigram 'the dog' that overlaps between {A} and {R}, and the total number of bigrams in {R} is $5$. The ROUGE-2 score can be calculated as the total number of bigrams in {A} that appear also in {R}, divided by the total number of bigrams in {R}. Thus, the ROUGE-2 scores, in this case, are $1/5$, whereas recall is $1/4$ and the F1-score is $0.22$ respectively. \\

    ROUGE-N is used only to calculate the error score over a single reference. However, when there are multiple references available then the ROUGE-N scores for each pair between {A} and every reference in reference set {R} is calculated. Finally, the maximum score is taken as the final score as follows:

\begin{equation}\label{eq:rouge-n}
    \text{ROUGE-N}_{multi} = \max_i \text{ROUGE-N}(r_i, {A})
\end{equation}
 
    Here, $r_i$ is the reference set, and {A} is the predicted response (i.e., candidate summary). \\

    {\bf ROUGE-L:} Is another version of ROUGE that uses the longest common subsequence (LCS) between {A} and {R}, where the matched words are not necessarily consecutive and the n-gram length is not required to be predefined. ROUGE-L can be calculated via two methods, named sentence-level LCS and summary-level LCS, to evaluate the generated response (i.e., summary). In the following, we discuss both approaches.

    Sentence-level LCS: This method treats the summary sentence as a sequence of words, where the similarity between {R} of length {m} and {A} of length {n} could be increased; if the matched LCS is found to be longer. The ROUGE-L score can be computed as follows:

\begin{equation}\label{eq:RougeRecall}
    R_{lcs}= \frac{LCS(A,R)}{m}
\end{equation}

\begin{equation}\label{eq:RougePrecision}
    P_{lcs}= \frac{LCS(A,R)}{n}
\end{equation}

\begin{equation}\label{eq:RougeF1}
    F_{lcs}= \frac{(1+\beta^2)R_{lcs}P_{lcs}}{R_{lcs}+\beta^2 P_{lcs}}
\end{equation}

    Here, LCS(A,R) is the length of longest common subsequence of predicted response {A} and given reference {R}, where $\beta$ is a non-negative real, so when $\beta = 1$, then $F_{lcs}$ is the harmonic mean of $R_{lcs}$ and $P_{lcs}$, and if $\beta < 1$, then ROUGE-L is weighted toward precision, otherwise, it is weighted toward recall. Thus, based on Equations \ref{eq:RougeRecall}, \ref{eq:RougePrecision} and \ref{eq:RougeF1} the ROUGE-L scores, recall, F1-score for Example \ref{example:4} are $3/5$, $3/6$ and $0.55$ respectively, with $\beta = 1$.

    Unlike bigram ROUGE-N which fails with capturing the different meanings among two different responses and the given reference, ROUGE-L can capture a  sentence-level structure in a natural way. However,  since LCS only counts the main in-sequence words, the other alternative LCS and shorter sequences can not be distinguished in the final score \cite{141}.

    Summary-level LCS: This measure can be thought of as a union of LCS matches between the reference $r_i$  and every candidate response {A}. The summary-level LCS score is calculated by  
    
\begin{equation}\label{eq:RougeRecall-s}
    R_{lcs}= \frac { \sum _{i=1}^{u}LCS_{\union}(A,R)}{m}
\end{equation}

\begin{equation}\label{eq:RougePrecision-s}
    P_{lcs}= \frac { \sum _{i=1}^{u}LCS_{\union}(A,R)}{n}
\end{equation}

\begin{equation}\label{eq:RougeF1-s}
    F_{lcs}= \frac{(1+\beta^2)R_{lcs}P_{lcs}}{R_{lcs}+\beta^2 P_{lcs}}
\end{equation}

\begin{example}
    Suppose R is given by "The quick brown fox jumps",  and the predicted response A1 is "The quick over the lazy"  and  A2 is "The brown lazy dog jumps".
\label{example:5}
\end{example}

    In this example,  the LCS of $R$ and $A1$ is "The quick", and the LCS of $R$ and $A2$ is "The brown jumps". The union LCS of $R$, $A1$, and of $R$, $A2$ is "The quick brown jumps" and the $LCS_{\union}(R,A) = 4/5$. \\

    {\bf ROUGE-W:} is the weighted longest common subsequence that considers the different spatial relations of LCS and their embeddings sequences.

\begin{example}
    Suppose that  the given R: 'The quick brown fox jumps over dog',  and the predicted response A1: 'The quick brown fox lazy so quickly'  and  A2: 'The lazy quick quickly brown dog fox'
\label{example:6}
\end{example}

    Here, $A1$ and $A2$ have the same ROUGE-L score, However, $A1$ is more adequate in terms of meaning than $A2$, because $A1$ has consecutive matches.

    To improve the basic LCS method and to give consecutive matches more awarded scores than non-consecutive matches, ROUGE-W introduces a weighting function that assigns different credit to consecutive in-sequence matches along with two-dimensional dynamic programming table $c(i,j)$ and $w(i,j)$, where $c$ is the dynamic programming table to store the weighted LCS (WLCS) score ending at words $a_i$  of $A$ and $r_j$ of $R$, and $w(i,j)$ is a table that store the length of consecutive matches ended at $c$ table position $i$ and $j$ \cite{141}. The weighting function must have the property that $f(A+R)> f(A)+ f(R)$ for any positive integers A and R. \\

    {\bf ROUGE-S:} is a skip-bigram concurrence score that allows to add a degree of indulgence to the n-gram matching that is performed with ROUGE-N and ROUGE-L. This allows any pair of consecutive words from $R$ that appear in $A$  to be considered even when the words are separated by any arbitrary gaps (i.e., one-or-more other words). The skip-bigram-based F-measure can be computed as follows:

\begin{equation}\label{eq:RougeRecall-sk}
    R_{skip2}= \frac { SKIP2(A,R)}{C(m,2)}
\end{equation}

\begin{equation}\label{eq:RougePrecision-sk}
    P_{skip2}= \frac { SKIP2(A,R)}{C(n,2)}
\end{equation}

\begin{equation}\label{eq:RougeF1-sk}
    F_{skip2}= \frac{(1+\beta^2)R_{skip2}P_{skip2}}{R_{skip2}+\beta^2 P_{skip2}}
\end{equation}

    Here, the $SKIP2(A,R)$ is the number of skip-bigram matches between $R$ and $A$,  $m$ is the length of $A$, $n$ is the length of $R$, $\beta\ $ is a parameter that control the relative importance of  $P_{skip2}$ and $R_{skip2}$, and C is the combination function.

\begin{example}
    Suppose R is: "the fox sat on the mat",  and the predicted response A is: "the white fox  and the dog"
\label{example:7}
\end{example}

    Here, the bigram "the fox" in $R$ would not match any bigram in $A$ if ROUGE-2 would be used since ROUGE-2 will try to find only exact consecutive bigram matches between $A$ and $R$. However, this is not the case with ROUGE-S, which will try to find any matched bigrams between $A$ and $R$ even when the words of the bigrams are separated by any arbitrary gap. In order to calculate the ROUGE-S score, first, we need to count the skip-bigram matches, SKIP2(A,R), within the maximum skip distance, so. In this example, {R} has the following skip-bigrams: ("the-fox", "the-jumps", "the-over", "the-dog", "fox-jumps", "fox-over", "fox-dog", "jumps-over", "jumps-dog", "over-dog"), and {A} has ("the-white", "the-fox", "the-and", "the-black", "the-dog", "white-fox", "white-and", "white-black", "white-dog", "fox-and", "fox-black", "fox-dog", "and-black", "and-dog", "black-dog"). Furthermore, there are three skip-bigram matches between {R} and {A} ("the-fox", "the-dog", "fox-dog"). Then by applying Equations \ref{eq:RougeRecall-sk}, \ref{eq:RougePrecision-sk} and \ref{eq:RougeF1-sk} one obtains the scores of $precision_{skip2}$, $recall{skip2}$ and $F{skip2}$ as $3/15=0.2$, $3/10=0.3$, and $(1+1) \times 0.3\times 0.2 / 0.3+1 \times 0.2)= 0.24$ respectively, with $\beta\ =1$.

\subsubsection{Metric for Evaluation of Translation with Explicit ORdering (METEOR)}

    Unlike BLEU, METEOR \cite{135} is based on the harmonic mean with a recall weighted higher than precision. METEOR is explicitly designed to work for sentence level rather than corpora level. METEOR  also has a set of variants (e.g., METEOR-NEXT, METEOR Universal, METEOR++, and METEOR++2) that came up with many features, such as: considering the word matching synonyms,  allowing relaxed matches with word stems, along with the standard exact word matching, applying importance weighting to different matching types,  allowing to integrate learned external paraphrasing resources (i.e., WordNet synonyms). METEOR correlates much better with human judgment, compared to BLEU, however, on contrary to BLEU which supports different lengths of n-grams, METEOR only considers unigram matches.

    In order to calculate the METEOR score, one needs first to form an alignment between {A} and {R}. An alignment is a set of mappings between the unigrams string, where each unigram in {A} must map {0 or 1} unigram in {R}. Then, the METEOR score computes the F-score using this relaxed matching strategy as follows:

First, precision and recall are calculated by
\begin{eqnarray}
    \text{Precision} = Pr = \frac{w} {\vert{w} \vert}
    \label{eq:precision} \\
    \text{Recall} = Re = \frac{w} {\vert{w}^{'} \vert}
    \label{eq:recall}
\end{eqnarray}

    where \(w\) is the number of mapped unigrams in {A} that are also found in {R}, and \(\vert{w}\vert\) is the total number of unigrams in {A} and \(\vert{w}^{'}\vert\) is the total number of unigrams in {R}.

    Second, the harmonic mean, $F_{mean}$, which places most of the weight on recall, is calculated by

\begin{equation}
    F_{mean}= \frac{ 10\times PrRe} {Re+9Pr}
\label{eq:fmean}
\end{equation}

    However, the longest contiguous unigram matches are rewarded by METEOR using a function named fragmentation penalty. Thus,  in order to find the longest sequence of adjacent unigrams that appear both in {A} and {R}, the unigrams need to be grouped into the fewest possible chunks of longest sequences with the fewest mappings. The fragmentation penalty is calculated as follows:

\begin{equation}
    Penalty= 0.5 (\frac{ c} {w_{n}}) ^{3}
\label{eq:penalty}
\end{equation}

    where {c} is the number of chunks and \({w_{n}}\) is the number of mapped unigrams. Here, the penalty increases as the number of chunks increases and vice versa.


    Finally, the METEOR score is calculated as follows:

\begin{equation}
    \text {\it METEOR: } score= F_{mean} (1-Penalty)
\label{eq:Meteor}
\end{equation}

    This can reduce the $F_{mean}$ score by a maximum of $50\%$ when there are no bigrams or longer matches found between {A} and {R}  \cite{135}.

    The procedure of calculating a METEOR score starts with finding the number of fragmentations with the longest sequence of adjacent unigrams that appear both in {A} and  {R}. Here, the number of possible fragmentations can be found as the number of possible chunks divided by the total number of unigram matches. Whereby, the longer unigram matches lead to a fewer number of chunks, and when there are no bigrams or longer matches, then the number of chunks is equal to the number of unigram matches, whereas the penalty is calculated according to Equation \ref{eq:penalty}.

    Let's consider the following examples to show how the METEOR score can be obtained.
\begin{example}
    Suppose {R} is given by "the dog sat on the mat" and {A} is "on the mat sat the dog".
\label{example:8}
\end{example}

\begin{example}
    Suppose {R} is given by "the dog sat on the mat" and {A} is "the dog was sat on the mat".
\label{example:9}
\end{example}

\begin{example}
    Suppose {R} is given by "the dog sat on the mat" and {A} is "the dog sat on the mat".
\label{example:10}
\end{example}

    For Example \ref{example:8}, we have $6$ chunks and $6$ unigram matches. Thus, the penalty is \(0.5 \times \frac{6}{6}=0.5\). The scores of precision, recall and F-mean are $1$,$1$, $1$ according  to Equations \ref{eq:precision},  \ref{eq:recall} and \ref{eq:fmean} respectively. Then, the final METEOR score can be calculated based on Equation \ref{eq:Meteor} as \(1\times(1-0.5)=0.5\).

    Following the same procedure, for Example \ref{example:10} we have $2$ chunks and $6$ matches which gives a METEOR score of $0.96$ and for Example \ref{example:10} we have $1$ chunk and $6$ matches with a final METEOR score of $0.99$.

\subsection{Machine-trained Evaluation Scores (MTES)}\label{sec:MTES}
	
    The measures in the category MTES are the most recent type of error scores. All of these measures have parameters that need to be learned from training data. This makes MTES more flexible than UAES. In the following, we discuss some of the most widely used scores in this category, i.e., ADEM \cite{150}, RUBER \cite{149}, RUSE \cite{182}, BLEURT \cite{151}, BertScore \cite{142}, MaUde \cite{153} and learning-based composite metrics \cite{181}.

\subsubsection{Automatic Dialogue Evaluation Model (ADEM)}

    Although many automatic evaluation metrics have been proposed to help reduce or eliminate the efforts of human evaluation in machine translation, the automatic evaluation for dialogue systems remains a challenge as traditional word-overlap metrics such as BLEU or ROUGE are considered as biased and do not correlate well with human judgment. Due to the fact that such metrics only detect the same words from the response and reference they are incapable of dealing with the semantic similarity of words. This problem might not be essential in evaluating machine translation as the range of words that could be used in the reference translation is rather limited, however, in a dialogue system the appropriateness of the spawned responses depends on the context and thus the influence can be more severe.

    In order to counter those drawbacks in the automatic evaluation of dialogue responses, learning-based scores were introduced which are models that can be trained through supervised, unsupervised, or semi-supervised learning to evaluate the score of a response.

    An example of such a parametric score is ADEM \cite{150}, which aims to learn distributed representations of the context, response and reference using RNN encoders through supervised learning.
    The goal of ADEM is to calculate a score given a context $c$, the reference $r$ and the model response $\hat{r}$, according to the following function:
\begin{equation}  
\begin{aligned}
\text{ADEM: } score(c,\hat{r},r; M, N) = (c^{T}M\hat{r}+r^{T}N\hat{r} - \alpha)/\beta
\end{aligned}\label{eqn.ademscore}
\end{equation}

    Overall, ADEM consists of two layers of RNNs. The first layer takes the input as a sequence of words and then produces a vector output at the end of each sentence (utterance). The second layer of the RNN takes each output utterance representation and produces a higher level of context representation from the last hidden state.

    The model is pretrained with VHRED \cite{183} which means the parameters of the RNN are frozen during the training, and the only trainable parameters are $M$,$N$ in Equation \ref{eqn.ademscore}. These two matrices project the model response $\hat{r}$ into the space of context and reference responses.
    By introducing $M$,$N$ into the model, the model will be able to give high scores for responses that have similar vector representations with the context and the reference responses.

    The training loss function $L$ of the ADEM model can be formulated as the squared error between the prediction ADEM score in Equation \ref{eqn.ademscore} and a human score plus an additional L2-regularization term:
\begin{equation}  
\begin{aligned}
L(M, N) =  \sum_{i=1:N}[score(c_i,\hat{r_i},r_i; M, N) - human\_score_i]^2 + \gamma\|\theta\|_2
\end{aligned}
\end{equation}
Here $\gamma\|\theta\|_2$ is the l2-regularization term where $\gamma$ is a scalar constant and $\theta$ is the set of parameters $\theta$ = \{M,N\}.

    A successful training will obtain the parameters of the model, $M$ and $N$, which minimize the loss $L$. One example of an ADEM score that correlates highly with a human score can be seen in the table \ref{tab:ADEMSCORE}.

	\begin{table}[t!]
		\caption{Examples for the comparison of scores for human score (HS), BLEU-2 (B2), ROUGE (RG) and ADEM (AM) (results from \cite{150}). All three scores are normalized to the same scale as human score.} 
		\centering 
		
		\begin{tabular}{p{3cm} p{1.7cm} p{2.8cm} p{0.3cm} p{0.3cm} p{0.3cm} p{0.5cm}} 
			\hline\hline 
			Context & Model responses & Reference responses & HS & B2 & RG & AM\\
			\hline
			i’d recommend \{url\} - or build buy an htpc and put \{url\} on it. → \textit{you’re the some nd person this week that’s recommended roku to me}. & 
			because it's brilliant &
			an htpc with xmbc is what i run . but i’ve decked out my setup. i’ve got \{number\} tb of data on my home server &
			5 & 1 & 1 & 4.726\\
			\hline
			imma be an auntie this weekend. I guess I have to go albany. here we go → \textit{u supposed to been here} → i come off nd on. → \textit{never tell me smh}& 
			haha, anyway, how're you?&
			lol you some timing&
			5 & 1 & 1 & 4.201\\
			\hline
			my son thinks she is plain. and the girl that plays her sister.seekhelp4him? → \textit{send him this. he’ll thank you. \{url\}} & 
			i will do &
			you are too kind for words &
			5 & 1 & 1 & 5.0\\
			\hline
		\end{tabular}
		\label{tab:ADEMSCORE}
	\end{table}

\subsubsection{Referenced metric and Unreferenced metric Blended Evaluation Routine (RUBER)}

RUBER \cite{149} is a learning-based hybrid score originally developed for the automatic evaluation of open-domain dialog systems. RUBER consists of a referenced part, $S_R$, and a unreferenced part, $S_{UR}$, hence, the term hybrid score. 

For the referenced part, $S_R$, the cosine similarity is evaluated between the generated representation of the response $\hat{r}$ and the representation of the ground truth response $r$ forming a score evaluating the similarity between the two responses. Both representations are obtained through pooling the maximum and minimal value from the word embeddings in each dimension given by:

\begin{equation}  
\begin{aligned}
    v_{max}[i] &=& maxpooling\{w_{1}[i],w_{2}[i],...,w_{n}[i]\} \\
    v_{min}[i] &=& minpooling\{w_{1}[i],w_{2}[i],...,w_{n}[i]\} 
\end{aligned}
\end{equation}

where $w_{1}[i],w_{2}[i],...,w_{n}$ are the words from the response, $i\in R^{d}$, $d$ is the dimension of the word embeddings. Then $v_{max}$ and $v_{min}$ are concatenated together by $v = v_{max} v_{min}$ to form the representation of reply.

The same procedure is applied to $\hat{r}$ and ground-truth reply $r$. Finally, the cosine distance is calculated to obtain the score which gives:

For the unreferenced score, $S_{UR}$, Bi-GRU RNN is utilized to measure the relatedness of the generated reply $r$ and its query $q$, the aim of this structure is to predict the quality of the generated reply given its query context. In order to do this the query and the reply are mapped to the word-level embeddings and then passed through two Bi-GRU RNNs individually where the last hidden states from forward and backward passes are concatenated to form the vector representations of $r$ and $q$. Finally $r$ and  $q$ form a quadratic feature, i.e., $q^TMr$, where $M$ is a parameter matrix are concatenated together and a multi-layer perceptron is applied to the final concatenated feature to predict a scalar score $S_UR$.

The objective of the model is to minimize the difference between the true pair $(q,r)$ and a randomly chosen false pair $(q,r^-)$ from the training set by at least a margin $\Delta$:

\begin{equation}  
\begin{aligned}
L = max\{0,\Delta-S_{UR}(q,r)+S_{UR}(q,r^-)\}
\end{aligned}
\end{equation}

The RUBER score can be used individually by either referenced or unreferenced part, however, the combined version of both parts by either minimal, maximum, geometric averaging, or arithmetic averaging provide the best RUBER score which is highly correlated with human annotation\cite{149}, a simple illustration of RUBER score can be seen in table\ref{tab:RUBERSCORE}.

\renewcommand{\arraystretch}{1.5} 
	\begin{table}[h]
		\caption{Examples for the comparison of scores for human score (HS), BLEU-2 (B2), ROUGE (RG) and RUBER unreferenced ($S_U$), RUBER referenced ($S_{UR}$) and RUBER Blended (RB) using the arithmetic mean (results are from \cite{149}). All four scores are normalized to the same scale as the human score (0-2).} 
		\centering 
		
		\begin{tabular}{p{0.8cm} p{1.4cm} p{0.4cm} p{0.4cm} p{0.4cm} p{0.4cm} p{0.4cm} p{0.4cm} p{0.4cm}} 
		
		
			
			\hline\hline 
			Query & True reply & Candidate reply & HS & B2 & RG & $S_U$ & $S_{UR}$ & RB\\
			\hline
			
			\multirow{2}{*}{It seems very near} &
			\multirow{2}{*}{Where are you} &
			I also think it's near &
			1.78 & 0.00 & 0.00 & 1.87 & 1.53 & 1.71\\

			                   &
			                   &
			Where are you from &
			1.78 & 0.00 & 0.77 & 1.15 & 1.78 & 1.47\\

			\hline
		\end{tabular}
		\label{tab:RUBERSCORE}
	\end{table}

\subsubsection{Metric for automatic Unreferenced dialogue evaluation (MaUdE)}

While the learning-based referenced metric ADEM shows good results over traditional word-overlap-based metrics, its nature of supervised learning requires human efforts for generating the score. This prevents ADEM from being applied to the field of online evaluation of dialogue systems where the generation of human reference scores is practically impossible. Furthermore, these types of referenced metrics have been shown to behave poorly on out-of-distribution data. In contrast, unreferenced metrics can easily overcome these challenges meanwhile keeping competitive performance against referenced metrics.

MaUde \cite{153} is a fully unreferenced metric developed for evaluating dialogue systems. It utilizes a pre-tained BERT model and applies a Noise Contrastive Estimation Technique \cite{187} to generate negative samples allowing completely unsupervised training. 
MaUdE provides a score for a given context \textit{c} and response \textit{r} by combining the learned representation vectors $f_e^{\theta_{1}}(c)$ and $f_e^{\theta_{2}}(r)$ using the combine function, $f_{comb}$. Then these are passed through a final classifier $f_c$ to produce a score $\{0,1\}$:
\begin{equation}  
\begin{aligned}
\text{MaUdE: } score(c,r) =     \sigma(f_c(f_{comb}(f_e^{\theta_{1}}(c),f_e^{\theta_{2}}(r)))
\end{aligned}
\end{equation}

where $f_e^{\theta_{1}}$ and  $f_e^{\theta_{2}}$ are the encoders to encoder the context and the responses respectively, and $c$ is the context that may contain many sentences $u_i$ and $c = \{u_1,u_2,u_3...u_n\}$. $f_e^{\theta_{1}}$ consist of a BERT model $f_e^{B}$ and a Bidirectional LSTM $f_e^{R}$. For obtaining representations of the context, each sentence $u_i$ passes through $f_e^{B}$ and then the encoded state $h_{u_i}$ is  passed through $f_e^{R}$. Finally, a max-pooling is applied to each hidden state $\hat{h_{u_i}}$ from $f_e^{R}$ and a weight matrix $W$ is applied to map the pooled representation to the response vector space for forming the final representation of context denoted by $h_{c_i}$. $f_e^{\theta_{2}}(r)$ was learned through the same procedure without the need of pooling operation to form a final representation of response $h_{r_i}$.

Finally, the context representation $h_{c_i}$ and the response representation $h_{r_i}$ are combined in the following way \cite{184}
\begin{equation}  
\begin{aligned}
f_{comb}(h_{c_i},h_{r_i}) =  concat([h_{c_i},h_{r_i},h_{c_i}*h_{r_i},h_{c_i} - h_{r_i}])
\end{aligned}
\end{equation}

The model is trained to distinguish between a correct response given by $score(c,r) \rightarrow 1$ and a false response corresponding to $score(c,r_n) \rightarrow 0$. 

A comparison of original MaUdE with baselines models trained with the same Noise Contrastive Estimation procedure, e.g., LSTM-based model unreferenced RUBER, InferSent and also BERT-based model DistilBERT-NLI \Comment{*unclear InferSent is no score, or is it?*}, showed that MaUdE obtained promising results, e.g., for the  PersonaChat dataset which is a public validation set. In addition, it was shown that MaUdE has a high correlation with human judgments on the PersonaChat dataset. Table \ref{tab:MAUDESCORE} demonstrates an evaluation example of a chat between a human and robot using several different models for evaluation.

\renewcommand{\arraystretch}{2} \Comment{Please remove this line of spacing argument if you find problems in other tables}
	\begin{table}[t!]
		\caption{Examples for a comparison of scores generated from unreferenced RUBER (R) based, InferSent (I) based, BERT-NLI (B) based baseline models against the original MaUde (M) model (results from \cite{151}). The conversation was generated between a human and a strong dialogue model (model that generated dialogues that have high human scores) \Comment{*what does this mean*}.} 
		\centering 
		
		\begin{tabular}{p{2.5cm} p{2.5cm} p{0.9cm} p{0.9cm} p{0.9cm} p{0.9cm}} 
			\hline\hline 
			Human & Model reply & RUBER & InferSent & BERT-NLI & MaUde\\
			\hline
			How are you today & 
			Hello how are you doing today
			& 0.943 & 0.908 & 0.998 & 1.0 \\

			Horrible, I'm about to file for divorce & 
			How long have you been married
			& 0.339 & 0.300 & 0.993  & 0.997 \\

			10 years, but she spends all my money! & 
			Wow that's a lot of money
			& 0.464 & 0.593 & 0.981 & 0.998 \\

			Yep. What do you do for a living & 
			I work as a scientist
			& 0.432 & 0.591 & 0.956 & 0.998 \\
			
			Do you have any hobbies & 
			I grow my farm
			& 0.247 & 0.284 & 0.980 & 0.999 \\
			
			So I guess you live in the country & 
			Yes I live in the country
			& 0.542 & 0.733 & 0.957 & 0.995 \\
			\hline
		\end{tabular}
		\label{tab:MAUDESCORE}
	\end{table}

\subsubsection{Regressor Using Sentence Embeddings (RUSE)}

RUSE \cite{182} is a learning-based score originally proposed for automatic evaluation for machine translation. 
RUSE formulates the training procedure as a regression problem that predicts the quality of a translation based on the machine translation hypothesis and the reference translation using human judgment data, therefore RUSE is a supervised model, in addition, RUSE also leverages the advantages of pre-trained universal sentence embeddings from InferSent \cite{184}, Quick-Thought \cite{185} and Universal Sentence Encoder \cite{186} by using the sentence embeddings produced by these models as the inputs, this critical step helps the model to better understand the semantic differences between sentences.

RUSE has two variants, one uses MLP (Multilayer Perceptron) as the regressor and the other one uses SVR (Support Vector Regression), though the author suggests that RUSE with MLP outperforms SVR version in many cases.

Once the sentence representations for MT hypothesis $t$ and the reference translation $r$ are generated using 3 universal sentence embeddings, they are concatenated to form a $d$ dimension vector $\hat{t}$ and $\hat{r}$, the final input feature to the regressor $f_r$ has $4d$ dimension and result from 3 types of matching methods:
\begin{eqnarray}  
\text{RUSE: } score = f_r(concat\{\hat{t},\hat{r}\},\hat{t}  *\hat{r},\|\hat{t} - \hat{r}\|)
\end{eqnarray}
where $\hat{t}*\hat{r}$ is the element-wise product and $\|\hat{t} - \hat{r}\|$ is the absolute element-wise difference.
The final regressor $f_r$ computes the score then the difference between the human judgment score and the produced score is measured to further adjust the parameters of the regressor during training.

The best setup of RUSE achieved the state-of-the-art result for the WMT16 and WMT17 datasets in both segment- and system-level metrics tasks\cite{149} by 2018.

\subsubsection{BLEURT}

This is a learning-based text generation score based on BERT. However, instead of using the naive BERT with a single pre-training procedure\cite{58}, BLEURT \cite{151} introduces another pre-training step on top of the basic pre-training to improve the generalization capabilities of the final score compared to most of the automatic scores even when the training data is very limited.

The training of BLEURT \cite{151} consists of 3 main steps: (I) Normal pre-training of BERT, (II) pre-training on large-scale synthetic data, and (III) fine-tuning on task-specific data. The key to the model is the second step which is the pre-training using large-scale synthetic data collected from Wikipedia. From this, a rich collection of synthetic reference-candidate pairs is generated by randomly perturbing the original sentences from Wikipedia. The perturbing technique includes mask-filling with BERT, back-translation, and dropping words. Then the original sentence $s$ and the generated pairs $\hat{s}$ are fed into BERT for 6 tasks which include regression tasks such as predicting BLEU, ROUGE, BERTscore, and the back-translation likelihood, and classification tasks such as textual entailment and back-translation flag prediction. It has been shown that the robustness of the model increases dramatically after the transfer learning phase on large-scale synthetic data.

The last step is to fine-tune a specific task and the goal of the model is to simply predict the score of a given sentence and the reference sentence. After all training steps, the model can be used to perform a human-like judgment score even when the dataset for fine-tuning is small and out-of-distribution\cite{151}.

\subsubsection{BERTScore}

BERTScore is an unsupervised learning-based metric proposed for automatic machine translation evaluation, it aims to calculate the semantic distance between the reference and the candidate sentence, by taking advantage of the unsupervised training of the BERT model, BERTScore offers better generalization to varies domains and is capable of understanding the semantic meaning of sentences especially capturing the distant dependencies and ordering information of sentences.

BERTScore computes the cosine distance between each token from reference sentence to each token candidate sentence using word embeddings produced by pre-trained BERT, the cosine similarity for a given reference token $x_i$ where $x = [x_1,x_2,...,x_n]$ and a candidate token $\hat{x_j}$ where $\hat{x} = [\hat{x_1},\hat{x_2},...,\hat{x_k}]$ of length $k$ is $cos = \frac{x_{i}^{T}\hat{x_{j}}}{||x_{i}||||\hat{x_{j}}||}$, the pre-normalized vectors are used in the calculation which leads to simply a inner product of $x_{i}^{T}\hat{x_{j}}$. The definition of BERTScore precision, BERTScore recall, and BERTScore F1 are:

\begin{equation}  
\begin{aligned}
\label{eq:BERTScore}
P_{BERT} = \frac{1}{\hat{x}}\displaystyle\sum_{\hat{x_{j}}\in \hat{x},x_{i}\in x}\max x_{i}^{T}\hat{x_{j}} \\
R_{BERT} = \frac{1}{x}\displaystyle\sum_{x_{i}\in x,\hat{x_{j}}\in \hat{x}}\max x_{i}^{T}\hat{x_{j}} \\
F_{BERT} = 2\times\frac{P_{BERT}\times R_{BERT}}{P_{BERT}+R_{BERT}}
\end{aligned}
\end{equation}

As several works discovered that rare words are more indicative for sentence similarity measure than common words\cite{145}, BERTScore involves Inverse Document Frequency (IDF) as a weight in the calculation, which further modifies the formula from \ref{eq:BERTScore} for example $R_{BERT}$ to:

\begin{equation}
    R_{BERT} = \frac{\sum_{x_{i}\in x}idf(x_i)\max_{\hat{x_{j}}\in\hat{x}}x_{i}^{T}\hat{x_{j}}}{\sum_{x_{i}\in x}idf(x_{i})}
\end{equation}

Since pre-normalized vectors are used in the calculation, $P_{BERT}$ and $R_{BERT}$ essentially have the range of the same cosine similarity of -1 and 1, in order to make the score more interpretable, $P_{BERT}$ and $R_{BERT}$ are rescaled to the range from 0 to 1.

Table \ref{tab:RUSEBLEURTBERTSCORE} shows some toy examples of 2 traditional word-overlap machine translation scores and 3 learning-based machine translation scores.

\renewcommand{\arraystretch}{2.5} 
	\begin{table}[t!]
		\caption{Examples for the comparison of scores mainly used for machine translation such as BLEU-2(B2), ROUGE-L(RO), RUSE with Infersent feature (RU), BLEURT (BL), and BERTScore-F1 (BE), all the sentence pairs are considered to have high semantic similarity by human annotators.} 
		\centering 
		
		\begin{tabular}{p{6.5cm}  p{0.6cm} p{0.6cm} p{0.6cm} p{0.6cm} p{0.45cm}} 
			\hline\hline 
			Reference ($x$) and candidate ($\hat{x}$) &  B2 & RO & RU & BL & BE\\
			\hline
			
			$x$: It is very cold today. \newline
			$\hat{x}$: The weather is freezing today. &
			0.00 & 0.40 & 0.42 & 0.62 & 0.76\\
			
			$x$: Wage gains have shown signs of picking up. \newline
			$\hat{x}$: Increases of wages showed signs of a recovery. &
			0.19 & 0.38 & 0.44 & 0.71 & 0.68\\

			$x$: What is your name? \newline
			$\hat{x}$: What can I call you? &
			0.00 & 0.22 & 0.10 & 0.48 & 0.41\\

			$x$: There are, indeed, multiple variables at play. \newline
			$\hat{x}$: In fact, several variables play a role. &
			0.00 & 0.28 & 0.40 & 0.75 & 0.56\\			
			
			$x$: Currently the majority of staff are men. \newline
			$\hat{x}$: At the moment the men predominate among the staff. &
			0.00 & 0.25 & 0.43 & 0.79 & 0.62\\		\hline	
		\end{tabular}
		\label{tab:RUSEBLEURTBERTSCORE}
	\end{table}

\subsubsection{Learning-based Composite Metrics} 

Automatic evaluation of image captioning is another challenging field. A good quality of an evaluation depends on accurate measures of both adequacy and fluency of the generated caption corresponding to lexical and semantic characteristics of a caption. However, most of the proposed metrics only focus on dealing with one of these two aspects of a caption. Learning-based Composite Metrics are proposed for the automatic evaluation aiming to address this problem by using a neural network in combination with different metrics taking advantage of each metric to produce a score that can effectively capture both adequacy and fluency of a caption.

The training of the Composite Metrics is formulated as a classification problem that distinguishes between a machine-generated caption and a human-generated caption using an MLP as a classifier. For each image, 3 machine-generated captions are produced by Show and Tell \cite{188}, Show Attend and Tell \cite{189} and adaptive attention \cite{190} are 3 well-known image captioning models respectively, and 5 human-generated captions will be used for each image. 
Each generated human caption is paired with the other 4 human-generated captions, for example, machine translation 1 for image $i$ will be paired with all human-generated captions for $i$ excluding caption 1, while machine translation 2 for image $i$ will be paired with all the human-generated captions for $i$ excluding caption 2. Each pair is considered to be one sample and each pair receives scores from METEOR, CIDEr, WMD, and SPICE. These scores are then used as input feature vectors for the classifier. Then the classifier predicts which samples are human-generated while machine-generated pairs are the negative samples, and human-generated pairs are the positive examples.

In this paper \cite{181}, it is stated that by combining different metrics and using them as input features to learn a classifier, the trained classifier can be further used to generate scores for automatic evaluation of image captions and this score is capable of capturing multiple aspects including adequacy and fluency. These results have been obtained by using the PASCAL-50s dataset.

\section{Challenges for evaluation scores} \label{sec.challenges}

From our discussion of quantitative error scores in Section \ref{sec.ces}, we can see that there is an impressive number of different error measures for evaluating QA systems. Regarding the historic development of such scores, it is interesting to note that till about 2014 there seem only $10$ automatic error measures that have been in wider usage, including BLEU, NIST and ROUGE \cite{139}. However, since then more than $50$ new scores have been introduced. An important contribution made by such new scores is that they add novel characteristics. Specifically, evaluation scores can have the following six properties:
\begin{itemize}
    \item word-based, character-based, n-gram-based or
word embeddings-based
    \item parametric or nonparametric
    \item context-dependent or context-independent
    \item composite measures or holistic measures
    \item human judgment-dependent or human
    judgement-independent
    \item transfer learning-based or non-transfer learning-based
\end{itemize}

It is important to note that the above properties lead to a certain self-similarity in the taxonomy of error scores. For instance, while all S-UAES and A-UAES do not involve human judgement, some scores in the category MTES do, e.g., ADEM or RUSE.

In the following, we discuss briefly the main drawback of particular properties of error scores.

\subsection{ Drawback of word-overlap scores}

A drawback of word-overlap scores (e.g., BLEU, ROUGE) is their inability to capture the semantic similarity between the answer and reference responses when there are few or even no common words. In \cite{150} it was argued that this problem is less severe for machine translation than dialog systems because for the latter the number of appropriate responses given in a context is much larger implying a response diversity. Indeed, almost all word-overlap scores were proposed for machine translation; and then adapted for other NLG tasks. However, in \cite{201} it was shown that even in machine translation word-overlap scores still have multiple weaknesses and cannot reflect the quality of a model fully.

Another drawback of word-overlap scores is not considering the context for the evaluation. Instead, they use only the response and reference. Importantly, in \cite{141} it was shown that neither of the word-overlap-based scores has any correlation to human judgment \cite{deriu2021survey}. For this reason, \cite{200} proposed to include human judgment into the BLEU score, which they called deltaBLEU. For deltaBLEU human judges rate the reference responses of the test set according to the relevance of the context. However, creating a high-quality reference dataset for this is a costly and arduous task. Furthermore, due to the fact that one question or query can have multiple correct answers, more than one reference needs to be created for each query.

\subsection{ Drawback of learning-based scores}

Despite the fact that learning-based scores constantly outperform word-overlap scores in evaluating semantic similarities,
there are some drawbacks. 
Some of the learning-based scores such as RUSE and ADEM are trained on data with human annotations, which is considered expensive as human annotation is time-consuming and costly. Due to this limitation, the training data is limited, hence they often suffer from significant performance loss when the test data is out-of-distribution.

Even though some later proposed metrics, e.g., BLEURT and BERTscore can utilize transfer learning to accommodate training data that cover more domains, the problem of domain shift still remains when the training data do not cover the entire scope of test data. Meanwhile, such metrics can only be used for evaluating a language that is trained for evaluation metrics, however, the evaluation quality across different languages can vary.


Another problem is that learning-based scores generally lack interpretability. This is related to the fact that human annotators usually assign several scores, e.g., according to different criteria (fluency, coherence, relevance, etc.). Furthermore, word-overlap scores are considered to capture only the syntactic similarity of a given context. Hence, people fail to explain what aspects the resulting score reflects. Instead, a general statement, for instance, "capturing semantic similarity" is provided by most authors to summarize the meaning of learning-based scores.

\subsection{ Drawback of scores based on a pre-trained model}

Usually, the training of a language model from scratch requires large amounts of data, which is not available in all situations. Hence, many scores are based on pre-trained models. However, the quality of the scores depends on the quality of the pre-trained models. Although the pre-trained models generally perform well, there is no guarantee that the performance is consistent in different situations, domains, and languages.

\subsection{Drawback of scores based on human judgement}\label{sec:HCES-DB}

In general, HCES based on human judgement is expensive, time-consuming, tedious, and requires domain expertise, which makes it cumbersome and impractical when dealing with a large task that may involve some dynamic behavior. Also, human judgments may differ from person-to-person, which reduces their reliability and reproducibility among different groups of experts \cite{5,144, 168}.

\subsection{Major problem}

We hypothesize that the major problem with current evaluations of QA systems can be formulated as follows:
\begin{quote}
The best way of evaluating QA systems is by human judgement. However, the problem is that such an evaluation is done qualitatively based on personal sentiments but the quantification of such a human judgement is still largely undefined. 
\end{quote}

Hence, the lack of a clear definition for the quantification of human judgement is the major obstacle to defining an MTCE and providing an automatic evaluation.

As a first step toward such a formulaization one could say the quantification of human judgement is based on a high-dimensional feature vector $v \in \mathbb{R}^d$ where $d\in \mathbb{N}$ is the number of features. As a second step, one could learn the representation of the vector space of $v$ via a deep learning neural network. This establishes the evaluation as a second model, while the first model is the QA system itself. In step three, we hypothezise that the evaluation model is at least as complex as the QA system. The reason for this is that it requires a complete comprehension of language semantics.

\section{Discussion}\label{sec.discussion}



In general, comparing two or more texts with each other is a non-trivial task. This is even further complicated by the fact that one can define dedicated sub-tasks thereof. Specifically, the problem of comparing two texts occurs in (I) machine translation, (II) text summarization, (III) question answering, (IV) dialogue systems (also called conversational agents), and (V) visual question answering. While each of these tasks produces a response in the form of text, the resulting forms of the output are quite different from each other. Put briefly, machine translation converts text automatically from one language into another language,  text summarization compresses a longer text into a shorter text without removing the semantic meaning of the longer text, question answering produces a response for a given question, and a dialogue system leads to an ongoing conversation and visual question answering describes information visually presented in form of an image or video. Due to the fact that each of the above sub-tasks is highly complex and an ongoing research field, we focus in this survey on question answering.

Starting from a discussion of the question answering framework (in Section \ref{sec.qaframework}) and its four main components, (i) QA algorithms, (ii), knowledge source, (iii) question types, and (vi) answer types, we introduced a formal definition of a question answering system (Section \ref{sec.defQAsystem}). This definition allows a summarization of practical elements of general QA systems by emphasizing their functional mappings. Hence, this abstract perspective enables an efficient focus on integral aspects of question answering. 

For the practical constructing of QA systems, we discussed the three main paradigms: (1) Information Retrieval-Based Question Answering (IRQA), (2) Knowledge Base Question Answering (KBQA), and (3) Generative Question Answering (GQA). Due to the fact that IRQA systems consist of two different components (retriever and reader), each with different functionalities, it is reasonable to use different scores to evaluate each component separately. The retriever component uses to retrieve the most candidate answers according to the {top-k} parameter that is chosen to determine the number of candidates to be returned. Evaluation scores such as Recall and MRR are then used to evaluate the retriever component and to test if the document that contains the right answer is among the retrieved candidates. Then the IRQA system directly ranks the answers retrieved from the knowledge source by considering information in the question itself. From the topic entity, the system may use different methods to extract all the entities and relationships related to a question, e.g., nodes and edges, and then arrange them, respectively. Next, the system encodes and analyzes the semantics of the input question and represents it as vectors. The semantic matching is then taken via vector-based measures to spread and aggregate the information along with the contiguous entities in the graph or within a given context. The answers are then arranged, and the higher-ranked entities are utilized to find the predicted answer for a given question. The reader component, in contrast, may use a simple score such as Exact Match or F1-score to evaluate the extracted (predicted) answer and find out to which extent it is close or matched with the gold answer. Such scores may capture the lexical overlap, but they cannot capture the similarity between two correct answers that differ in their semantic content. Some examples for Exact Match and F1-score are given in Section \ref{sec: S-UAES}. However, more complex error scores like BERTScore or BLEURT are needed in order to capture semantic similarity.

Predicting an answer in a KBQA system depends on the ability of reading-comprehension that is responsible for fully absorbing and understanding the question \cite{119, 120}. This is achieved by conducting a semantic and grammatical analysis that is required for obtaining the encoded question, and then representing it logically in a justified manner according to grammatical rules defined by the system. The logical form can be validated by a semantic alignment for the structured KB or by simultaneously performing a logical analysis with knowledge base grounding. The logical models are then validated in the KB during partial parsing. Thus, the logic model of a KB is analyzed and the predicted answers are extracted.

It is interesting to note that some IRQA systems have also been designed to perform complex reasoning on the question structure and flexible semantic matching that may fit easily into the nature of end-to-end training. However, the interphases of this system may be less interpretable. In general, KBQA can provide good insights and explanations by developing a logical form, but it still relies heavily on the logical form design and the parsing algorithm, which has been proven to be a bottleneck in improving performance. Due to the complexity of knowledge representation, mapping user questions to logical or semantic queries, and providing an answer, is not an easy task for KBQA, especially for complex questions.

In contrast, a GQA system can be considered an open-ended text-generation problem that may not include all possible references related to a particular question. This implies that for a semantic evaluation one would not only need human annotated references but many alternative forms thereof to capture the many variations which can be generated by a GQA system itself. This is an unsolved problem because of the associated costs and complexity. Therefore, GQA systems are frequently not evaluated semantically but lexically, e.g., using n-gram overlap, for UAES (such as Precision, Recall, F1, BLEU, ROUGE, METEOR) \cite{114}. However, it should be clear that computing the n-gram matches between the generated question and reference does usually not allow to distinguish between  salient or important from non-salient or non-important types of n-grams. Hence, such scores are not able to weigh different n-grams based on their importance or based on the actual content, instead, they weigh all  n-grams equally, and this indifference can affect the performance negatively.

In general, drawing a conclusion about the quality of a QA system requires an informed assessment using annotated datasets together with evaluation scores that allow the quantification of the performance. For this reason, many datasets with different structures (e.g., document-based or knowledge-based) have been prepared for different question types (e.g., standalone, sequential, multiple-choice) along with different answer types (i.e., simple, abstractive, agnostic) for benchmarking QA systems (see Section \ref{sec.benchdata}). The different benchmark data address the different requirements and variations of  QA tasks. It is interesting to note that since the beginnings question-answering datasets have evolved dramatically becoming more and more complex. This acknowledges also the requirements of GQA systems due to their flexibility to generate text in an open-ended form (see discussion above).

The heterogeneous tasks of QA systems demand not only problem-specific benchmark data but also heterogeneous evaluation scores. This explains the large number of different evaluation scores that have been proposed over the years. In Section \ref{sec.errorscores}, we introduced a hierarchical taxonomy showing that such error scores can be broadly classified into human-centric evaluation scores (HCES) and automatic evaluation scores (AES). In general, HCES is considered the best evaluation that gives the most trustworthy score, however, due to drawbacks mentioned in Section \ref{sec:HCES-DB}, AES are more practical. ASE measures are further sub-categorized in Untrained Automatic Evaluation Scores (UAES) and Machine-trained Evaluation Scores (MTES) to address the requirements of the agnostic or specific tasks of a QA system.

The UAES measures do not require any pre-training, because such measures do not depend on parameters. In contrast, MTES contain adjustable components, i.e., parameters, which need to be estimated for a given task via a learnable model. The MTES measures can be further sub-categorized in composite and holistic measures. The composite measures are also called feature-based measures, and holistic measures are called end-to-end measures. The difference is that a holistic measure cannot be separated into individual error scores while this is the case for feature-based measures where the features correspond to such individual error scores.

Finally, we would like to highlight that the availability of many evaluation scores posses the problem of selecting an appropriate one \cite{152}. This is another non-trivial task because the interpretation and selection of the presented evaluation scores is challenging and currently there is no universal score that would be best in all situations. For instance, a score such as F1 may show adequate results on some existing spanning-based datasets, but it may be inadequate for other QA tasks. Hence, for each situation a comparative analysis is needed, similar to model selection \cite{ding2018model,emmert2019evaluation}, for identifying the most favorable evaluation score.


	\section{Conclusion}\label{sec.conclusion}
	
    The question-answering task is among the oldest challenges in artificial intelligence and it is still one of the most important tasks in natural language processing to this day as it enables humans to interact with a machine in a natural way. For dealing with domain-specific information and different formats of data which could be either structured or unstructured data there are three main paradigms for constructing QA systems: (1) Information Retrieval-Based Question Answering, (2) Knowledge Base Question Answering, and (3) Generative Question Answering.

    In this paper, a comprehensive survey of question-answering systems was presented based on the general architecture of the question-answering framework. Despite the heterogeneity of QA tasks and design principles, we identified commonalities and introduced a general definition of a question-answering system allowing us to summarize its formal structure abstractly. In order to understand key assumptions and design principles behind QA systems, this paper described also task-specific benchmark datasets and evaluation scores. For the latter, taxonomy has been introduced where its main branches are separating human-centric evaluation scores (HCES)from automatic evaluation scores (AES) and untrained automatic evaluation scores (UAES) from machine-trained evaluation scores (MTES). Overall, we have seen that not only the construction of a QA system is complex but also its evaluation and an open problem is the quantitative formalization of the human judgment underlying HCES which could be translated into MTES.


	\bibliographystyle{unsrtnat}
	\bibliography{QA}

\end{document}